\documentclass{article}
\usepackage[final]{nips_2017}
\usepackage[utf8]{inputenc} 
\usepackage[T1]{fontenc}    
\usepackage{hyperref}       
\newcommand{\doi}[1]{doi: \href{https://doi.org/#1}{#1}}
\usepackage{url}            
\usepackage{booktabs}       
\usepackage{amsfonts}       
\usepackage{amsmath}
\usepackage{nicefrac}       
\usepackage{microtype}      
\usepackage{graphicx}
\usepackage{listings}
\usepackage{xcolor}
\usepackage{pgfplots}
\usepackage{tikz}
\usepackage{subcaption}
\usepackage{multirow}
\usepackage{cellspace}
\usepackage{amssymb}

\usepackage{array}

\usepackage{ragged2e}

\usepackage{placeins}
\usepackage{float}

\usepackage{caption}
\pgfplotsset{compat=1.18}

\makeatletter
\renewcommand{\@notice}{}
\renewcommand{\paragraph}{%
  \@startsection{paragraph}{4}{\z@}%
                {1.5ex \@plus 0.5ex \@minus 0.2ex}%
                {0.5ex \@plus 0.2ex}%
                {\normalsize\bf}}
\makeatother

\title{Safe Reinforcement Learning for Multi-Turn Clinical Agents: When Dense Rewards Beat Outcome Rewards}

\author{
  Brendan P. Murphy \\
  \texttt{cs.brendan@gmail.com} \\
  \And
  Agustina Saenz, MD, MPH \\
  \texttt{asaenz@stanford.edu} \\
}

\begin{document}
\maketitle
\thispagestyle{plain}

\begin{abstract}
Reinforcement learning with verifiable rewards is increasingly used to improve language model reasoning, yet its effects on sequential decision procedures remain poorly understood. We study this problem in clinical multi-turn agents that must decide what information to request, when to stop gathering evidence, and which action to take. We introduce three controlled environments: medication refill triage, acute diagnostic work-up, and dosing titration. Each environment includes hidden clinical states, query costs, exact adaptive oracles, and deterministic outcome verifiers. Across these tasks, outcome-only group relative policy optimization often improves aggregate reward while producing undesirable procedures, including premature commitment, one-question collapse, and failures on rare high-risk cases. In diagnostics, assigning dense credit to questions that reduce uncertainty improves held-out value and moderate-case support across three training seeds relative to outcome-only reinforcement learning, without the high-risk seed collapse observed under outcome-only training. Separating exploration from exploitation further increases high-risk support at its best checkpoint, but at the cost of excessive information gathering on low-acuity cases. Additional experiments show that the optimal strength of process credit is task-dependent; after a supervised warmup that already gathers carefully, reinforcement learning raises aggregate reward by trimming questions while degrading high-risk performance; and reinforcement learning does not reliably repair rare behavior gaps inherited from biased supervised data. These findings show that verifiable final outcomes are insufficient for training reliable multi-turn agents. Reward design must account for the procedure used to reach an answer, and evaluation must distinguish adaptive information gathering from reward-exploiting shortcuts. More broadly, the results identify reward geometry, initialization coverage, and task headroom as interacting determinants of whether reinforcement learning improves or degrades sequential reasoning in safety-critical settings.

\end{abstract}

\section{Introduction}
\label{sec:intro}

Clinical decision making is often a sequential process of gathering evidence under
uncertainty. A clinician evaluating acute chest pain may need to obtain an
electrocardiogram, cardiac biomarkers, haemodynamic findings, and symptom history
before deciding whether emergency treatment is warranted. A clinician reviewing a
medication refill may need recent renal function, adverse effects, adherence, and
contraindications before approving or modifying therapy. A clinician adjusting an
anticoagulant must balance clotting risk against bleeding risk, often using information
that is incomplete at the beginning of the encounter. In each setting, the quality of
the decision depends not only on the final action, but also on which information was
requested, how that evidence was interpreted, and when the decision maker chose to
stop gathering.

Clinical language agents face the same sequential problem. An agent that acts from
an incomplete prompt may miss a dangerous condition; an agent that asks every
available question may create unnecessary delay and burden. The desired policy must
therefore learn three related behaviours: what to ask, when enough information has
been obtained, and which action to take. This makes clinical multi-turn interaction
different from one-shot medical question answering. The target is not merely a
correct response to a fully specified vignette; it is a reliable decision procedure
for reaching an action from an initially incomplete state.

Recent research has begun to evaluate medical language models as
sequential decision makers rather than static question-answering
systems. Nori et al.\ introduce the Sequential Diagnosis Benchmark
(SDBench), in which agents iteratively request clinical findings and
diagnostic tests before committing to a diagnosis
\citep{nori2025sequential}. Their MAI Diagnostic Orchestrator (MAI-DxO)
demonstrates that structured, cost-aware inference-time orchestration
can improve diagnostic efficiency without updating model parameters.
This motivates a complementary learning question: whether reinforcement
learning can induce effective information-gathering procedures within
the policy itself, and under what conditions those procedures improve
or degrade.

Reinforcement learning with verifiable rewards, or RLVR, is a natural candidate for
training such procedures. Supervised fine-tuning can teach an action schema,
clinical protocols, and examples of desirable conversations. It does not directly
optimize the tradeoff between requesting more evidence and committing to a decision.
Reinforcement learning can, in principle, optimize that tradeoff. If the final action
can be scored, and each additional question carries a cost, an agent can be rewarded
for gathering useful information while avoiding unnecessary investigation. Recent
work has extended RLVR from mathematics and code to medical reasoning
\citep{chen2024huatuogpto1, zhang2025medrlvr, lin2025ehrmind}; however, most
medical RLVR evaluations still emphasize the correctness of a final answer rather
than the quality of the sequential process used to obtain it.

Many language agents now trained with reinforcement learning share a gather-then-commit
structure: search agents decide what to retrieve before answering, coding agents what to
inspect before editing, and tool-use agents which calls to make before acting, yet the
dominant recipe rewards only the final outcome. Whether outcome-only training teaches a
sound information-gathering procedure or a reward-exploiting shortcut is therefore a
general open question, and clinical decision-making is its high-stakes instance.

This distinction creates an important safety problem. A terminal reward may assign
a high score to the correct clinical action even when that action was reached through
an unreliable procedure. A model may commit after one question, choose a broadly
conservative action that performs adequately on common cases, or avoid a difficult
decision by selecting an intermediate option. Such policies can improve average
reward while becoming less reliable on rare, high-risk cases. Conversely, a reward
that strongly encourages information gathering may cause the model to ask too many
questions, delay action, or escalate unnecessarily. Aggregate reward alone may
therefore conceal clinically important changes in behaviour.

We study these learning dynamics in three controlled clinical environments:
medication refill triage, acute diagnostic work-up, and dosing titration. Each
environment is formulated as an ask-to-commit decision process. The agent begins
with an incomplete case, may request individual findings, receives deterministic
answers from a simulator, and eventually commits to a clinical action. The
environment includes a hidden clinical state, an explicit cost for each question,
a hand-specified utility function, and an exact adaptive oracle that identifies the
optimal sequence of questions and actions. Cases are generated from the simulator
rather than extracted from clinical records. The environments are therefore not
intended as clinical validation benchmarks; their purpose is to isolate how
reinforcement learning changes information gathering when the state, reward, and
optimal policy can be measured exactly.

\begin{figure}[p]
    \centering
    \includegraphics[width=\textwidth]{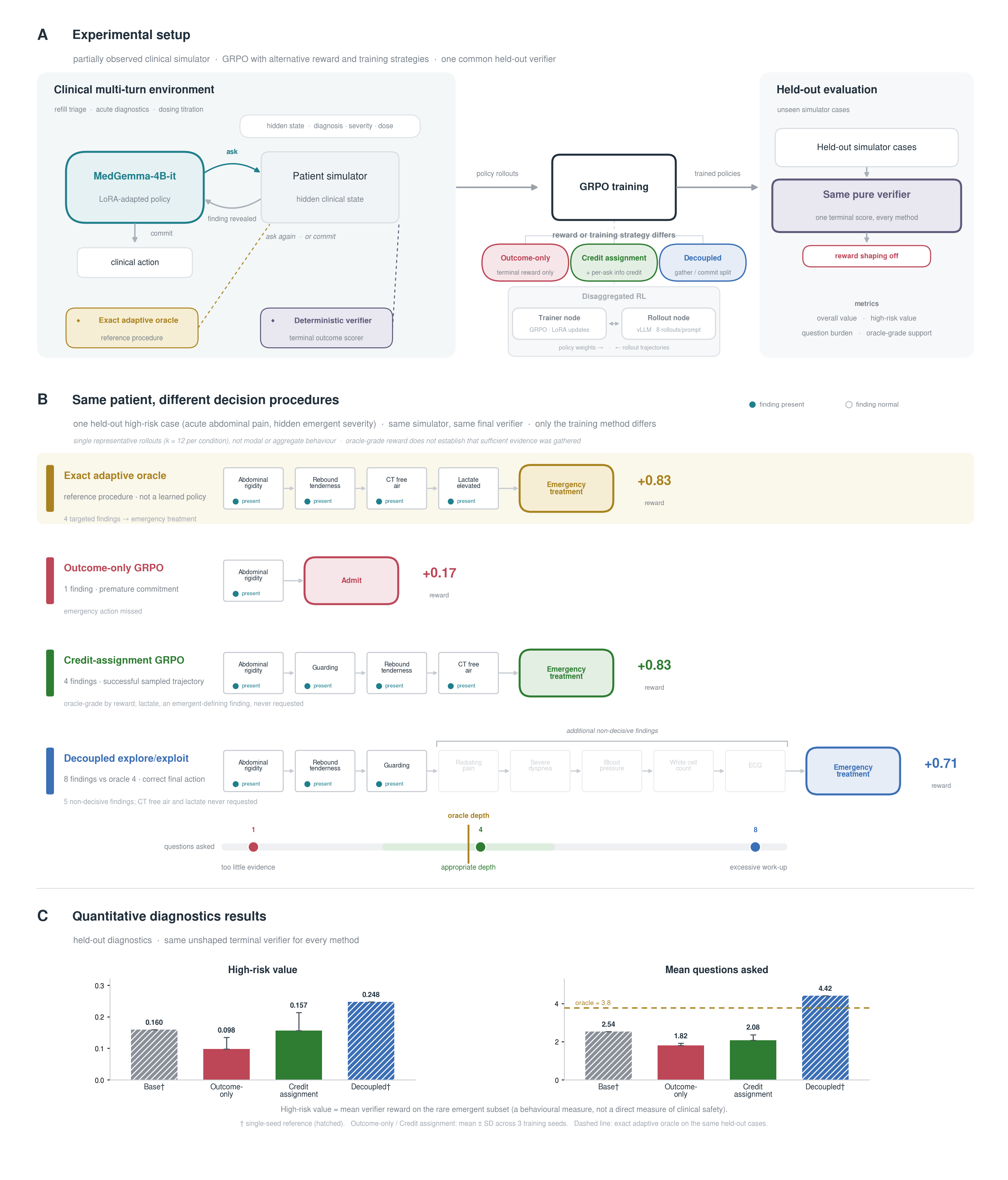}

\caption{
\textbf{Experimental framework and effects of reward design on clinical
multi-turn decision procedures.}
\textbf{A,} The LoRA-adapted MedGemma-4B-it policy interacts with a
partially observed patient simulator across medication-refill triage,
acute diagnostics and dosing titration. Each environment provides an
exact adaptive oracle and deterministic outcome verifier. The three
GRPO methods use different training reward formulations or exploration
strategies, with policy optimization and rollout generation distributed
across two compute nodes. All policies are evaluated on held-out cases
using the same unshaped terminal verifier.
\textbf{B,} Selected trajectories for the same simulated high-risk
abdominal-pain case illustrate different information-gathering procedures.
The exact adaptive oracle gathers four targeted findings before emergency
treatment. Outcome-only GRPO commits prematurely after one finding.
Credit-assignment GRPO reaches the correct terminal action after four
requests and receives an oracle-grade reward despite following a different
evidence sequence. Decoupled exploration reaches the correct terminal
action after eight requests, including additional non-decisive findings.
The learned trajectories are selected illustrative examples, not
representative estimates of typical policy behavior. Oracle-grade reward
does not necessarily establish sufficient gathered clinical evidence.
\textbf{C,} Held-out diagnostics performance measured by high-risk
verifier value and mean questions asked. Outcome-only and
credit-assignment GRPO show mean $\pm$ standard deviation across three
training seeds; the base model and decoupled GRPO are single-seed
references (hatched bars). The dashed line indicates the exact adaptive
oracle's mean question count. All methods are scored using the same
unshaped evaluation reward. Decoupled GRPO is shown at its step-40
checkpoint, selected on held-out data; its final checkpoint does not
retain the high-risk advantage (Table~\ref{tab:dx}). The held-out
diagnostics set contains only two high-risk cases.
}
\label{fig:overview}

\end{figure}

Figure~\ref{fig:overview} summarizes the sequential decision problem and
illustrates how different reward designs can produce qualitatively different
information-gathering procedures for the same underlying clinical state.

The central question is whether RL learns a better decision procedure, or merely a
more effective way to exploit the reward. We first compare outcome-only group
relative policy optimization, or GRPO, with credit-assignment GRPO, which gives
additional training credit to questions that reduce uncertainty about the hidden
clinical state. We then study a decoupled exploration and exploitation variant that
separates information gathering from final commitment. Finally, we examine how the
benefit of RL depends on supervised initialization, including a setting in which
supervised fine-tuning already performs strongly and a setting in which the
supervised data contain a rare high-risk coverage gap.

The results show that a verifiable terminal reward is not sufficient to produce a
reliable multi-turn policy. Plain GRPO often increases average reward by shortening
the interaction, collapsing toward one-question policies, or choosing actions that
perform adequately on common cases while failing rare emergencies. In the
diagnostics task, dense credit for uncertainty-reducing questions improves held-out
value and moderate-case support across three training seeds, without the high-risk seed
collapse seen under outcome-only GRPO. Separating exploration from exploitation produces
the strongest high-risk support at its best checkpoint, but it also increases
information gathering on low-acuity cases. The dosing experiments show
that the strength of process credit acts as a task-dependent control parameter; too
little credit permits under-gathering, while too much encourages excessive
gathering. The supervised initialization experiments reveal a further limitation:
after a supervised warmup that already gathers carefully, RL raises aggregate reward
by trimming questions while degrading high-risk performance, and it does not reliably repair a rare high-risk coverage gap even when
the missing action remains possible to sample.

The primary conclusion is therefore not that RL reliably improves clinical agents.
Rather, RL can improve, preserve, or degrade the information gathering procedure
depending on the task, initialization, and reward structure. Adaptive headroom is
required because asking questions must offer an advantage over a fixed action.
Useful behaviour must also be represented in the initial policy because policy
optimization cannot reliably amplify trajectories that are never generated.
Finally, the reward must assign credit to the procedure itself; otherwise, outcome
optimization can favour shortcuts that score well without gathering the evidence
needed for safe decisions.

These findings have implications for future clinical decision support systems.
A model should not be considered safer merely because its average verifier score
improves. Evaluation should also examine whether the model obtained the findings
needed to justify its action, whether it committed prematurely, whether it
overinvestigated low-risk cases, and whether performance was preserved on uncommon
but consequential scenarios. In real clinical systems, these concerns would extend
to incomplete records, noisy patient responses, clinician disagreement, workflow
burden, and uncertainty that cannot be represented by a deterministic verifier.
The present environments simplify those factors, but they expose a basic risk that
will remain relevant in more realistic systems: reinforcement learning can alter the
decision procedure in clinically important ways even when final answer metrics appear
to improve.

We make three main contributions. First, we introduce a controlled framework for
studying clinical multi-turn agents as ask-to-commit decision processes with exact
verifiers, exact adaptive oracles, and separate evaluation of high-risk cases.
Second, we show that outcome-only GRPO can learn shortcut procedures even when the
task has measurable value of information and the initial model can sample useful
behaviour; dense credit assignment improves and stabilizes the main diagnostics
result. Third, we characterize several limits of RL post-training: stronger
exploration trades high-risk performance against low-acuity burden, process credit
must be calibrated to the task, strong supervised models may leave little safe
headroom, and RL does not reliably repair coverage gaps inherited from biased
supervised data.

More broadly, this work frames reinforcement learning for clinical multi-turn agents
as procedure learning rather than final answer optimization. Safe post-training
requires more than a verifiable outcome; it requires a task in which adaptive
gathering has value, an initialization that exposes useful behaviour, a reward that
assigns appropriate credit to intermediate decisions, and an evaluation that audits
rare failures rather than relying on aggregate performance alone.

\FloatBarrier
\section{Background}
\label{sec:background}

\paragraph{Verifiable reward learning for reasoning and medicine}
Reinforcement learning with verifiable rewards has been most successful when the
quality of an output can be checked directly. In mathematics, a proposed answer
can be compared with a known solution; in programming, generated code can be
executed against tests \citep{shao2024deepseekmath}. Recent work has extended
this approach to medical reasoning by rewarding answers that agree with
references, structured rubrics, or rule-based evaluators
\citep{chen2024huatuogpto1, zhang2025medrlvr, lin2025ehrmind}. These studies
suggest that verifiable feedback can improve medical reasoning, but they largely
treat the model response as a completed product. Many clinical decisions instead
emerge through a sequence of information requests, interpretations, and actions.
For these problems, the object being optimised is not only the final answer; it
is the procedure used to reach that answer.

This distinction is especially important in clinical settings. A correct
medication decision may depend on whether renal function, allergies, adverse
effects, or interacting therapies were checked first. An appropriate emergency
disposition may depend on whether the agent identified and obtained the findings
needed to justify escalation. A verifier that scores only the terminal action
can therefore reward a policy that reaches the correct answer without following
a procedure that would remain reliable when the case changes. Verifiability
makes the learning signal measurable, but it does not by itself make the learned
behaviour clinically appropriate.

\paragraph{Relation to RLHF and preference-based clinical alignment}
Not every clinically important property can be represented by a
deterministic verifier. Communication quality, uncertainty expression,
shared decision making, and preference-sensitive tradeoffs may require
human judgement. Clinical post-training has therefore also explored
preference-based alignment. Savage et al.\ compare supervised
fine-tuning with direct preference optimisation across clinical tasks
\citep{savage2025clinicaldpo}, while Dou et al.\ use diagnostic-process
feedback to construct preferences for multi-turn medical dialogue
\citep{dou2024physicianlogic}. These approaches complement the
analytically specified reward setting studied here.

We study a complementary part of clinical alignment. In our environments, the
hidden state, the cost of requesting information, and the utility of the final
action can be specified analytically. This permits exact comparison with an
adaptive oracle and makes it possible to identify whether reinforcement
learning improves the decision procedure or exploits the scoring rule. We use
RLVR because it provides this experimental control, not because deterministic
rewards can represent every aspect of clinical quality. The distinction also
matters when interpreting learned process supervision: feedback from another
language model is a learned judgement signal, rather than a strictly verifiable
reward, and it may introduce the preferences or biases of the supervising
model.

The broader safety concern applies to both settings. If a policy can exploit a
clean analytic verifier, it may also exploit a learned reward model. Preference-based clinical alignment should therefore be evaluated not only through average
reward or final answer accuracy, but also through audits of premature
commitment, unsupported escalation, unnecessary information gathering, and
performance on uncommon high-consequence cases.

\paragraph{Clinical multi-turn agents}
Clinical multi-turn agents interact with patient states that are incomplete at
the beginning of an encounter. They must choose which question to ask next,
update their decision after receiving the answer, and determine when further
information is unlikely to change the appropriate action. This differs from
medical question answering, where the relevant findings are usually presented
in a complete vignette and the model produces a single response.

Recent medical dialogue agent research supports this sequential framing. For
example, DoctorAgent RL uses a doctor agent that interacts with a language model
patient agent and receives feedback for diagnostic accuracy, information
acquisition, and protocol compliance \citep{feng2026doctoragentrl}. Conversational
diagnostic systems such as AMIE \citep{tu2024amie} elicit history through dialogue, and
ED-Copilot learns cost-aware laboratory test selection for emergency department
triage \citep{sun2024edcopilot}. Such work
moves medical language models closer to proactive consultation, where the agent
must elicit missing evidence rather than wait for a fully specified case. Our
focus is narrower and more controlled. We use deterministic simulators to isolate
how reinforcement learning changes three parts of the policy: which information
the agent requests, when it stops requesting information, and which action it
takes after gathering evidence.

SDBench and MAI-DxO provide a complementary approach to sequential
clinical reasoning \citep{nori2025sequential}. SDBench evaluates
diagnostic accuracy and investigation costs as agents acquire clinical
information over multiple turns, while MAI-DxO uses inference-time
orchestration to structure hypothesis refinement and test selection.
Unlike this prompted orchestration approach, we investigate how
reinforcement learning changes the information-gathering policy itself.
Our controlled environments expose exact hidden-state dynamics,
deterministic outcome verifiers, and adaptive oracle policies, enabling
direct measurement of premature commitment, unnecessary investigation,
and high-risk failures. These simplifications permit a more controlled
study of learning dynamics, although they do not reproduce the clinical
realism of SDBench.
AgentClinic is an interactive clinical simulation benchmark in which
language-model agents gather information from simulated patients and clinical
tools before making diagnostic decisions \citep{schmidgall2026agentclinic}.
Whereas AgentClinic evaluates clinical agents in interactive environments, we use
controlled, deterministic environments with exact adaptive oracles to study how
RL reward design changes information-gathering behaviour.

This controlled setting represents a simplified analogue of several realistic
clinical workflows. In refill triage, an agent may need to check symptoms,
adherence, recent laboratory values, and contraindications before continuing or
modifying treatment. In acute assessment, it may need to obtain red-flag
findings before deciding whether emergency intervention is warranted. In dosing
titration, it may need to balance competing treatment risks before increasing,
reducing, or withholding therapy. These examples motivate the decision
structure, although the synthetic environments do not model the noise,
ambiguity, workflow constraints, or clinician disagreement present in actual
care.

\paragraph{Reinforcement learning shapes procedures, not only answers}
Post-training can alter how a language model allocates inference time and
organises intermediate reasoning, even when evaluation ultimately concerns a
final output. SPIRAL, for example, trains a model to coordinate search, multiple
reasoning traces, and aggregation, thereby optimising a broader inference
procedure rather than a single generated answer \citep{hamid2026spiral}. Our
setting studies a different procedure: clinical gather-then-commit behaviour.
The model must request relevant findings, use the revealed evidence, and commit
only when additional information is no longer worth its cost.

This procedural view changes how performance should be interpreted. Two agents
may choose the same final action but differ substantially in reliability. One
may have gathered the findings that support the decision; the other may have
guessed correctly after a single question. Similarly, an agent that asks every
possible question may eventually choose the correct action while imposing
unnecessary delay and burden. Final action accuracy alone cannot distinguish
these policies. Evaluation must also measure how the action was reached.

\paragraph{Why dense credit matters for clinical information gathering}
A terminal outcome reward provides feedback only after the dialogue has ended.
When a model asks several questions before committing, that reward does not
identify which question changed the decision, which question reduced
uncertainty, or which question was unnecessary. Plain GRPO therefore faces a
credit assignment problem. A successful trajectory may contain useful and
wasted actions, yet every token in the trajectory is updated from the same final
outcome. This can encourage simpler shortcuts, particularly when asking fewer
questions reduces the explicit interaction cost.

Prior work on mathematical reasoning has distinguished final-answer
correctness from the correctness of intermediate steps. Uesato et al.\
investigated process- and outcome-based feedback
\citep{uesato2022processoutcome}, followed by work on human-labelled
process supervision \citep{lightman2023verify} and automatically
constructed process feedback for verification and reinforcement learning
\citep{wang2024mathshepherd}. Related approaches to multi-step optimization include OREO, which jointly learns a policy and value function from offline trajectories
\citep{wang2025oreo}, and SWiRL, which decomposes tool-use
trajectories into action-level subtrajectories optimized using learned
feedback \citep{goldie2025swirl}. On-policy self-distillation instead
provides token-level teacher supervision on student-generated
trajectories \citep{zhao2026selfdistilledreasoner}.

Our approach instead assigns analytic information-gain credit to
observable clinical information requests, without requiring a learned
reward model or supervision of free-form reasoning traces. This credit
is accumulated within the training reward for on-policy GRPO over
complete dialogues, rather than used to optimize individual
subtrajectories.

The process credit is used only during training. Every policy is evaluated with
the same terminal utility and question cost, without the additional shaping
signal. Any improvement must therefore appear as better gather-then-commit
behaviour under the original task objective. This separation between training
credit and evaluation reward is central to the experimental design. It allows us
to test whether dense credit teaches a more reliable procedure, rather than
merely increasing the shaped reward that the model was trained to optimise.

\section{Experimental setup}
\label{sec:setup}

We study clinical multi-turn agents in a controlled ask-to-commit setting. Each
episode begins with an incomplete clinical case. The agent may request missing
findings, receive answers from a deterministic simulator, and then commit to one
clinical action. Requesting information carries a small cost; the agent must
therefore balance the value of additional evidence against delay and unnecessary
questioning. The desired behaviour is not to ask as little or as much as possible.
It is to obtain the information needed to support an appropriate action, then stop.

This design captures a simplified version of several clinical workflows. A refill
agent may need recent renal function or adverse effect history before continuing a
medication. An acute assessment agent may need red-flag findings before choosing
emergency treatment. A titration agent may need to compare treatment benefit with
bleeding or toxicity risk before changing a dose. The environments remove much of
the ambiguity of real care, but preserve the sequential decision problem that the
experiments are intended to study.

Figure~\ref{fig:overview} summarizes the three clinical tasks, the supervised and
reinforcement learning stages, the reward variants, and the common held-out
evaluation procedure.

\paragraph{Environment family}
All three tasks instantiate the same partially observed decision process. A hidden
clinical state, denoted by $\mu$, represents the underlying patient condition. In
refill triage and diagnostics, $\mu$ is a latent severity level. In titration, it
is a joint clot and bleed risk state. Each case also contains a vector of binary
clinical findings. Their values are sampled from task-specific prevalences, then
held fixed throughout the episode. The agent observes only findings that are
initially provided or later requested.

At each turn, the policy emits one structured JSON action. It may either
\emph{ask} about a single finding or \emph{commit} to a terminal clinical action.
An ask is answered by the simulator; a commit ends the episode. Episodes are
limited by a task-specific turn budget, which prevents unlimited questioning and
makes failure to reach a decision explicit.

Each question has cost $c{=}0.03$. The terminal reward is

\[
R = U[\mu][a] - c\,n_{\text{asks}},
\]

where $U$ is a hand-specified utility matrix, $a$ is the committed action, and
$n_{\text{asks}}$ is the number of questions requested. The utility matrix assigns
different consequences to each action under each hidden state. A decision that is
reasonable for a low-risk patient may therefore receive a poor score when the true
state is high-risk.

The question cost makes the clinical tradeoff explicit. For example, if the
committed action has utility $0.90$ and the agent asks three questions, the final
reward is $0.90 - 3(0.03) = 0.81$. Asking an additional useful question is
worthwhile only when the information is expected to improve the final decision by
more than its cost. Asking a repeated or irrelevant question lowers reward without
changing the patient state.

\paragraph{Exact adaptive oracle}
Because the simulator, state prior, finding probabilities, question cost, and
utility matrix are known, we can compute an exact adaptive oracle for every task.
The oracle chooses the next question by considering how each possible answer would
change the optimal action. It stops when no available question has enough expected
value to justify its cost.

This oracle provides a target for the complete decision procedure, not only the
terminal action. It identifies which findings should be requested, when gathering
should stop, and which action should follow from the revealed evidence. A sampled
trajectory is considered \emph{oracle-grade} when its realised reward is within
$\text{EPS}{=}0.03$ of the oracle value. The tolerance equals the cost of one question, so a policy may differ from the oracle by one marginal request without
being treated as a substantively different procedure.

\paragraph{Clinical tasks}
We instantiate the framework in three stylized tasks that represent distinct forms
of clinical uncertainty.

\emph{Refill triage} models a medication continuation decision. The agent may need
to ask about renal decline, dehydration, adverse effects, contraindications, or
other medication safety findings before approving, modifying, denying, or
escalating a refill. The two medication families are ACE inhibitor or angiotensin
receptor blocker therapy and nonsteroidal anti-inflammatory drug therapy.

\emph{Acute diagnostics} models the work-up of chest or abdominal presentations.
The agent must decide whether to continue routine evaluation, choose an
intermediate disposition, or initiate an emergency action. The task is deliberately
constructed so that the emergency action requires several supporting findings.
This exposes a clinically important shortcut: a policy can perform adequately on
common cases by choosing an intermediate action while failing uncommon emergencies.

\emph{Dosing titration} models an ordered treatment decision under competing risks.
The agent chooses among three treatment intensities while considering both clot and
bleed risk. Too little treatment may leave the patient unprotected; too much may
increase harm. The correct action therefore depends on the combination of findings,
rather than on a single severity variable.

\begin{table}[ht]
\centering
\caption{
Clinical environment dimensions. ``Findings'' is the set of binary clinical
features available for the agent to request. ``Commit actions'' is the number of
terminal decisions available to the policy. All environments use a question cost
of $0.03$ and an oracle-grade tolerance of $0.03$. Diagnostics contains the
largest finding space and is designed to reveal hedging and premature commitment
on rare emergency cases.
}
\label{tab:envs}

\resizebox{\textwidth}{!}{%
\begin{tabular}{lccccc}
\toprule
Task & Findings & Commit actions & Families & Latent state & Turn cap \\
\midrule
Refill triage & 12 & 5 & 2 (ACEi/ARB, NSAID) &
severity (none/mod/high) & 10 \\
Diagnostics & 18 & 5 & 2 (chest, abdominal) &
severity (none/mod/high) & 12 \\
Titration & 6 & 3 (ordered) & 1 (anticoagulation) &
joint clot$\times$bleed (9 states) & 10 \\
\bottomrule
\end{tabular}%
}
\end{table}

\paragraph{Case generation and data splits}
The core experiments do not use an external clinical dataset. Instead, cases are
generated on demand from the task simulators. Each generated case contains a
latent state, a fixed set of finding values, a clinical family, and the associated
oracle policy. This approach provides exact labels and complete control over case
prevalence; it does not reproduce the noise, missingness, or demographic complexity
of clinical records.

RL training uses $300$ generated cases per task. Cases are drawn from the hidden
condition, $\rho{=}0$, in which no decision-relevant findings are revealed at the
start of the episode. Training cases use generation seed $12345$. Evaluation cases
use seed $100$ and are generated independently, so the training and held-out sets
are non-overlapping samples from the same simulator distribution.

The main held-out evaluation contains $n{=}60$ cases in the hidden condition. For
the supervised coverage experiment, we increase the evaluation set to $n{=}200$ so
that the uncommon high-risk subgroup contains enough cases to support a separate
analysis. The SFT warmups use $800$ oracle or checklist trajectories generated
with seed $7$. These trajectories demonstrate the action format and desirable
question sequences before reinforcement learning begins.

\paragraph{Training rollouts and evaluation sampling}
During GRPO training, the policy generates a group of eight rollouts for each
prompt. Relative rewards within that group determine the policy update. The model
therefore learns from comparisons among several candidate ask and commit
trajectories generated from the same initial case.

During held-out evaluation, we sample $k{=}12$ trajectories per case.
Following the coverage-based perspective of repeated sampling
\citep{brown2024monkeys}, we define support@$k$ as the fraction of cases
for which at least one sampled trajectory is oracle-grade. This
metric measures the availability of successful behaviour within the
policy distribution, rather than its reliability in a single
deployment trajectory. We therefore interpret support alongside
realised value, mean number of questions, risk subgroup results,
and audits of the trajectories themselves.

Unless otherwise stated, experiments use training seed $1$. The main comparison
between plain GRPO and credit-assignment GRPO on diagnostics is repeated across
three training seeds. The remaining experimental cells use one seed and are treated
as exploratory results; this limitation is discussed in
Section~\ref{sec:discussion}.

\paragraph{Safety-focused evaluation}
Aggregate reward can conceal clinically important failures because high-risk cases are
rare: the held-out sets contain 2 of 60 high-risk cases for diagnostics, 6 of 60
high-clot cases for titration, 7 of 60 high-risk cases for refill triage, and 14 of 200
for the supervised coverage experiment. We therefore report that subgroup separately throughout the paper. We also examine the
number and type of questions asked, premature commitment, one-question collapse,
unsupported escalation, hedging, and excessive gathering.

This distinction is necessary because two policies can receive similar aggregate
scores for very different reasons. One may gather the findings required to support
its action. Another may reach the same action through a fixed bias or a lucky guess.
Likewise, a policy may improve high-risk performance by escalating broadly, while
creating unnecessary intervention on benign cases. The evaluation is designed to
separate these behaviours rather than treating every high scoring trajectory as
equally desirable.

Representative cases and oracle conversations for all three tasks are provided in
Appendix~\ref{app:cases}. Full SFT and GRPO hyperparameters are reported in
Appendix~\ref{app:impl}.

\paragraph{Scope and clinical interpretation}
The environments are synthetic, stylized, and labelled by deterministic rules.
They are not clinical validation benchmarks and should not be interpreted as
evidence that any trained policy is suitable for patient care. The utility matrices
are experimental design choices, not estimates of real clinical outcomes. Patient
answers are deterministic, findings are binary, and the simulator does not model
documentation errors, ambiguous symptoms, conflicting records, patient
preferences, or clinician disagreement.

These simplifications are intentional. By controlling the hidden state, available
questions, question cost, utility function, and oracle policy, we can identify how
reinforcement learning changes the information gathering procedure. The setup
therefore supports conclusions about learning dynamics and reward design; it does
not support conclusions about clinical efficacy. Its relevance to future decision
support lies in the failure modes it exposes: premature action, unnecessary
questioning, unsupported escalation, and apparently improved average performance
that conceals deterioration on rare high-consequence cases.

\section{Methods}
\label{sec:method}

The experiments follow a common sequence. We first test whether adaptive information
gathering can improve the task and whether the initial policy can produce useful
trajectories. We then compare outcome-only GRPO with two methods that change how
credit is assigned to information gathering. Finally, we evaluate every trained
policy using the same unshaped verifier. This separation ensures that a method
cannot appear successful merely because its training reward is easier to optimise.

\subsection{Pre-RL eligibility gates}
\label{subsec:gates}

Before running policy optimisation, we apply two inexpensive eligibility tests. The
first measures whether the task contains meaningful value of information. The
second measures whether the initial policy can sample the behaviour that RL is
intended to reinforce. These gates do not guarantee that RL will succeed; rather,
they identify conditions in which RL has little plausible opportunity to help.

\paragraph{Gate 1: adaptive headroom}
The first gate compares the exact adaptive oracle with the best policy that commits
without requesting any information. We define the resulting best constant gap as

\[
\text{BC gap}
=
V_{\text{oracle}}
-
\max_a \mathbb{E}_{\mu}\!\left[U[\mu][a]\right],
\]

where $V_{\text{oracle}}$ is the expected value of the optimal adaptive policy and
$U[\mu][a]$ is the utility of action $a$ under hidden state $\mu$.

A large positive gap indicates that asking questions can improve the decision over
the best fixed action. A gap near zero indicates that the task offers little
adaptive headroom; even an ideal information gathering policy performs only
slightly better than committing immediately. In that setting, a small RL gain may
reflect the structure of the task rather than a failure of the optimisation
algorithm.

\paragraph{Gate 2: latent behavioural support}
The second gate tests whether the initial policy can sample an oracle-grade
trajectory under stochastic decoding. For every held-out case, we generate
$k{=}12$ independent rollouts. A case is counted as supported when at least one
rollout reaches a realised reward within $\text{EPS}{=}0.03$ of the exact adaptive
oracle. The proportion of supported cases defines support@$k$.

This gate measures whether useful behaviour exists somewhere in the initial policy
distribution. If support is zero for the target behaviour, policy gradient
optimisation has no successful trajectory to amplify. In that situation, additional
supervised training or improved data coverage is more appropriate than beginning
RL directly from the current model.

Support@$12$ should not be interpreted as ordinary deployment reliability. It is
a capability and exploration measure; a policy may produce one successful rollout
among twelve while behaving poorly on most individual trajectories. We therefore
interpret this gate separately from greedy performance, realised value, question
burden, and high-risk subgroup behaviour.

Gate 1 is analytic and depends only on the environment. Gate 2 is an inference-only
probe of the initial policy, either the base model or an SFT checkpoint. Neither
gate uses the training rewards introduced below.

\subsection{Model and optimisation setup}
\label{subsec:model-setup}

All methods use the text-only \texttt{medgemma-4b-it} model
\citep{saab2025medgemma}. We adapt the model using LoRA
\citep{hu2022lora}, which limits the number of trainable parameters while preserving
the same base checkpoint across comparisons. Reinforcement learning uses group
relative policy optimisation, or GRPO \citep{shao2024deepseekmath}. The implementation
uses PyTorch \citep{paszke2019pytorch}, Hugging Face Transformers
\citep{wolf2020transformers}, and the GRPO trainer from TRL \citep{vonwerra2020trl}.

Training is fully on policy. For each clinical prompt, the model generates a group
of candidate multi-turn trajectories, receives a reward for each trajectory, and
is updated according to relative performance within the group. The same model
produces both information requests and terminal actions.

We use a disaggregated two-node system. One node performs optimisation, while a
separate vLLM server generates rollouts from the current policy
\citep{kwon2023vllm}. Simulator responses
are inserted between model turns but are excluded from the policy gradient using
an \texttt{env\_mask}. Consequently, gradients are applied only to tokens generated
by MedGemma; the model is not trained to imitate the deterministic patient
simulator.

\subsection{Outcome-only GRPO control}
\label{subsec:plain-grpo}

Outcome-only GRPO is the principal control. The policy receives the terminal
clinical utility minus the accumulated cost of requesting information:

\[
R_{\text{terminal}}
=
U[\mu][a] - c\,n_{\text{asks}}.
\]

No intermediate question receives a separate reward. The entire trajectory is
therefore updated from the final action, the true hidden state, and the total number
of questions asked.

This is the most direct use of a clinical outcome verifier. It rewards an
appropriate action and discourages unnecessary questioning. It also creates a
credit assignment problem. If a trajectory contains several information requests,
the model is not told which request changed the decision, which was irrelevant,
or whether the final action succeeded despite inadequate evidence.

Outcome-only optimisation can therefore favour procedures that are inexpensive but
unreliable. A policy may commit after one question, repeatedly choose an
intermediate action that performs reasonably on common cases, or escalate broadly
without gathering the findings needed to justify the decision. These behaviours
can improve average reward while weakening the intended gather-then-commit
procedure.

\subsection{Credit-assignment GRPO}
\label{subsec:credit-assignment}

Credit-assignment GRPO adds a dense training signal for questions that reduce
uncertainty about the hidden clinical state. The terminal clinical objective is
unchanged; the additional signal is intended to help the model identify which
information requests contributed to the decision.

After each newly revealed finding, we update the posterior distribution over the
hidden state. The information credit at turn $t$ is the corresponding reduction
in posterior entropy:

\[
\Phi_t
=
H\!\left(
p(\mu \mid \text{revealed}_{t-1})
\right)
-
H\!\left(
p(\mu \mid \text{revealed}_{t})
\right).
\]

The training reward becomes

\[
R_{\text{train}}
=
U[\mu][a]
-
c\,n_{\text{asks}}
+
\lambda \sum_t \Phi_t,
\]

where $\lambda$ controls the relative strength of information credit.

A question receives positive credit when its answer reduces uncertainty about the
latent state. A repeated question, or a question that does not change the
posterior, receives no information credit. The coefficient $\lambda$ therefore
controls how strongly the model values informative questions relative to their
explicit cost.

This construction is motivated by potential-based reward shaping
\citep{ng1999shaping}; however, we use it as an auxiliary process reward and do not
claim that it is policy invariant under every task and training configuration.
The coefficient can alter the preferred procedure. A weak coefficient may fail to
prevent under-gathering, while a strong coefficient may reward excessive
questioning.

The distinction between training and evaluation is central. Information credit is
used only during rollout training. It is not added to the utility matrix, the
environment score, either pre-RL gate, or any reported evaluation reward. A credit-assignment policy must therefore improve the actual gather-then-commit trajectory
to perform better under evaluation.

\subsection{Decoupled exploration and exploitation GRPO}
\label{subsec:decoupled}

Credit-assignment GRPO still optimises asking and committing within a single-trajectory. The two behaviours may compete for the same gradient signal; asking
more can increase information credit, while committing earlier can reduce question
cost. To test whether a stronger separation improves high-risk behaviour, we
implement a decoupled exploration and exploitation variant inspired by DREAM
\citep{liu2021dream}.

This method is not an exact implementation of DREAM. It is a single-model adaptation designed for the present ask-to-commit environments. The model uses one
network and one LoRA adapter, with a mode token indicating whether the current
update concerns exploration or exploitation.

During the exploration phase, the policy may request findings but may not commit.
It is rewarded for gathering evidence that makes the hidden state recoverable.
During the exploitation phase, the policy selects a terminal action using only the
findings gathered during exploration. Its reward is the pure clinical utility of
that action.

The two phases receive separate gradient updates, so the information gathering
signal and the commitment signal are normalised within their respective phases.
At evaluation time, the phases are composed into one dialogue. The policy first
gathers information, then commits using the resulting transcript.

A hard information boundary prevents the exploitation phase from observing the
privileged latent state or any finding that the exploration phase did not request.
This guard is necessary to ensure that improved terminal decisions arise from
gathered evidence rather than hidden simulator information.

The method tests whether stronger exploration improves rare high-risk decisions.
It also exposes the opposite failure mode. A policy that is strongly rewarded for
recoverability may gather more information than is clinically necessary,
particularly in benign or low-acuity cases.

\subsection{SFT followed by RL}
\label{subsec:sft-rl}

Practical post-training pipelines generally apply supervised fine-tuning before
reinforcement learning. In this setting, the relevant question is not whether RL
can teach the task from the base model; it is whether RL improves behaviour that
the supervised model has not already learned.

We study two forms of SFT initialization. In the refill triage task, RL begins from
a light protocol-teaching warmup. The supervised trajectories demonstrate the
structured action format and common medication safety procedures. This experiment
tests whether RL offers additional value after SFT has already learned much of the
target behaviour.

In the diagnostics task, we construct a coverage-biased SFT checkpoint by reducing
the proportion of high-risk oracle trajectories in the supervised data. This
creates an initialization with strong aggregate performance but a specific rare
case weakness. We then measure whether the policy can still sample the missing high-risk behaviour and whether credit assignment or decoupled GRPO can repair the
coverage gap.

These experiments separate two limitations. A well-covered SFT policy may leave
little adaptive headroom for RL; the supervised model already performs strongly on
the measured objective. A biased SFT policy creates a different problem; the
desired behaviour may be present only weakly or rarely in the policy distribution.
Passing the latent support gate makes repair possible in principle, but does not
guarantee that RL will preferentially reinforce the missing behaviour.

\subsection{Common evaluation and guarded reporting}
\label{subsec:evaluation}

We evaluate all trained policies on held-out cases using the pure terminal verifier:

\[
R_{\text{eval}}
=
U[\mu][a] - c\,n_{\text{asks}}.
\]

No information credit, exploration reward, or supervised trajectory score is used
during evaluation. This common objective permits direct comparison among methods
that were trained with different reward structures.

We report overall realised value, oracle-grade support@$k$, and the mean number of
questions requested. Where an experiment is replicated, we report variation across
training seeds. We also analyse the high-risk subgroup separately because uncommon
cases can contribute little to the mean while accounting for the most consequential
failures.

Aggregate metrics are not sufficient to determine whether the procedure improved.
We therefore use a guarded report that audits the structure of each trajectory. The
report includes:

\begin{itemize}
    \item \textbf{Blind commitment:} the policy commits without requesting any
    decision-relevant information.

    \item \textbf{One-question collapse:} the policy repeatedly commits after a
    single request, even when the oracle requires additional evidence.

    \item \textbf{Premature commitment:} the policy acts before obtaining the
    findings needed to support the decision.

    \item \textbf{Unsupported escalation:} the policy selects a high-acuity action
    without gathering the evidence required by the task.

    \item \textbf{Hedging:} the policy overuses an intermediate action that performs
    adequately on common cases but avoids the correct high-stakes decision.

    \item \textbf{Over-gathering:} the policy requests substantially more
    information than the oracle, particularly on benign or low-acuity cases.

    \item \textbf{High-risk regression:} performance deteriorates on the high-risk
    subgroup even when aggregate value improves.
\end{itemize}

A policy may therefore receive a higher mean reward and still be classified as
learning an exploitative procedure. For example, broadly selecting emergency
treatment may reduce missed emergencies but does not constitute safer information
gathering when the escalation is unsupported and benign cases are also over-treated. Similarly, support@$12$ may remain high when one successful rollout is
embedded within a distribution dominated by premature decisions.

The central evaluation criterion is whether the policy improves the complete
procedure: it should gather the findings needed for the decision, avoid irrelevant
requests, commit at an appropriate point, and preserve performance on rare high-risk cases. Final outcome reward remains important, but it is interpreted together
with the evidence gathering behaviour that produced it.

\section{Results}
\label{sec:results}

\begin{figure}[p]
\centering
\includegraphics[width=\textwidth]{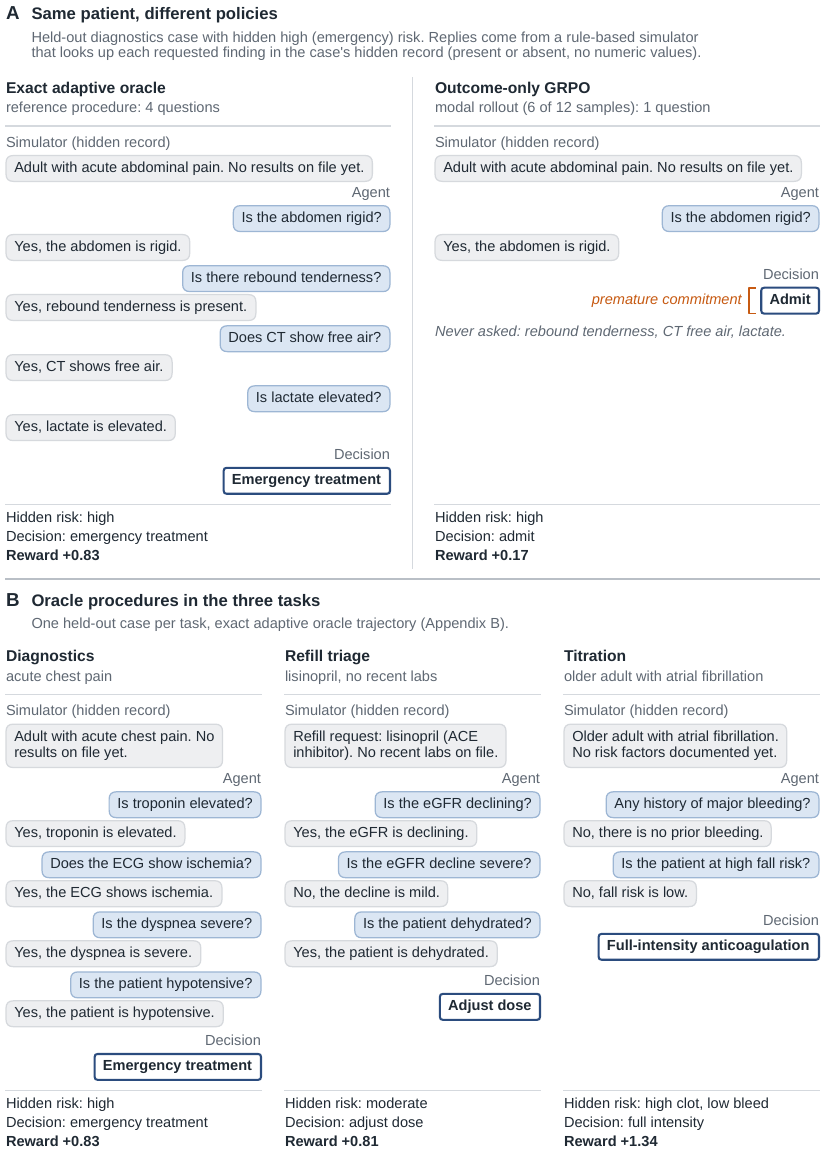}
\caption{Agent--simulator dialogues rendered in plain language. The dialogues are
simplified simulator traces, not real clinical conversations. The replying party is
a rule-based simulator, not a patient or clinician. It looks up each requested
finding in the case's hidden record and reports whether it is present. Findings
are binary, so the simulator returns no numeric values; each question and answer
paraphrases one requested finding and its revealed value.
\textbf{(A)}~The same held-out high-risk diagnostics case as Figure~\ref{fig:overview}.
The exact adaptive oracle gathers four findings before emergency treatment (reward
$+0.83$). The outcome-only GRPO rollout shown is that policy's most common behaviour
on this case (6 of 12 samples). It asks one question and admits the patient (reward
$+0.17$).
\textbf{(B)}~Exact adaptive oracle trajectories for one held-out case in each task,
matching Appendix~\ref{app:cases}. Rewards are terminal utility minus $0.03$ per
question. Hidden risk is never shown to the agent.}
\label{fig:dialogues}
\end{figure}

Figure~\ref{fig:dialogues} shows what these procedures look like as conversations.
The experiments reveal a consistent distinction between improving a terminal
score and learning a reliable information gathering procedure. Outcome-only GRPO
can increase aggregate value while reducing the amount of evidence gathered and
weakening performance on rare high-risk cases. Credit assignment improves the
main diagnostics comparison, but its effectiveness depends on the task and the
strength of the information reward. Stronger separation between gathering and
commitment improves high-risk support, but increases questioning on low-acuity
cases. The SFT experiments show a further constraint; RL adds little safe improvement
when the supervised policy has already captured the available behaviour, and it does not
reliably repair a rare coverage gap inherited from supervised training.

We present the results in six stages. First, we show that passing the pre-RL gates
does not ensure that outcome-only GRPO will learn a safe procedure. Second, we
compare outcome-only GRPO with credit assignment across three diagnostics seeds.
Third, we test whether separating exploration from exploitation improves high-risk
behaviour. Fourth, we examine how the information credit coefficient controls
under-gathering and over-gathering in titration. Fifth, we measure the remaining
RL headroom after an SFT warmup. Finally, we test whether RL can repair a high-risk
coverage gap in a biased SFT policy.

Table~\ref{tab:master} summarizes the final held-out results. High-risk value and
high-risk oracle-grade support are the primary safety-focused metrics. Overall
value, overall support, and mean questions remain important, but none is
sufficient in isolation. In several experiments, an apparently favourable
aggregate score is produced by premature commitment, aggressive action selection,
or excessive questioning rather than by adaptive evidence gathering.

\begin{table}[ht]
\centering

\caption{Final held-out results across tasks and training methods. All policies
are evaluated in the hidden condition using the pure verifiable reward and
$k{=}12$ sampled rollouts per case. High-risk value and high-risk oracle-grade
support are the primary safety-focused metrics. Overall value, overall support,
and mean questions describe aggregate performance and interaction burden. Bold
marks the strongest non-exploitative high-risk result within each task. Apparent
support produced by a degenerate under-gathering policy is marked
\textsc{exploit}, because support@$12$ can remain high even when typical
trajectories do not gather the evidence required for the decision.
$^\ddagger$Diagnostics plain GRPO and credit-assignment GRPO are reported as
means across three seeds; the corresponding standard deviations appear in
Table~\ref{tab:dx}. All other rows are single-seed results.
High-risk subgroup sizes are small (diagnostics $n{=}2$, titration high-clot $n{=}6$,
refill $n{=}7$ cases); with $k{=}12$ rollouts per case, diagnostics high-risk support
can take only the values $0$, $0.5$, and $1$.}
\label{tab:master}

\resizebox{\textwidth}{!}{%
\begin{tabular}{llccccc}
\toprule
Task & Method & overall val & overall supp & high val & high supp & asks \\
\midrule
\multirow{4}{*}{Diagnostics}
 & base                 & $+0.251$ & $0.35$ & $+0.160$ & $0.50$ & $2.5$ \\
  & plain GRPO$^\ddagger$ & $+0.285$ & $0.41$ & $+0.098$ & $0.33$ & $1.8$ \\
 & credit-assignment GRPO$^\ddagger$ & $+0.313$ & $0.59$ & $+0.157$ & $\mathbf{0.50}$ & $2.1$ \\
 & decoupled GRPO (step 40$^\S$) & $+0.329$ & $0.42$ & $\mathbf{+0.248}$ & $\mathbf{1.00}$ & $4.4$ \\
\midrule
\multirow{3}{*}{Titration}
 & base                 & $-0.022$ & $0.52$ & $+0.396$ & $0.67$ & $2.8$ \\
 & plain GRPO           & $-0.012$ & $0.57$ & $+0.627$ & $1.00^\dagger$ & $1.4$ \\
 & credit-assignment GRPO ($\lambda{=}0.1$) & $-0.036$ & $0.52$ & $\mathbf{+0.556}$ & $\mathbf{1.00}$ & $2.7$ \\
\midrule
\multirow{4}{*}{Refill triage}
 & base                 & $+0.386$ & $0.58$ & $+0.151$ & $0.00$ & $2.2$ \\
 & SFT warmup           & $+0.318$ & $0.97$ & $\mathbf{+0.195}$ & $1.00$ & $2.8$ \\
 & plain GRPO           & $+0.431$ & $0.97$ & $+0.032$ & $1.00^\dagger$ & $1.3$ \\
 & credit-assignment GRPO            & $+0.415$ & $0.93$ & $+0.043$ & $1.00^\dagger$ & $1.6$ \\
\bottomrule
\end{tabular}%
}

\vspace{2pt}
{\footnotesize
$^\dagger$ High-risk support is a $k$-fold sampling artifact over a degenerate
under-gathering policy (mean asks $\ll$ oracle), not adaptive gathering; the
guarded report labels these \textsc{exploit}.\\
$^\ddagger$ Diagnostics plain GRPO and credit-assignment GRPO are 3-seed means;
full mean$\pm$std values are reported in Table~\ref{tab:dx}. All other rows are
single-seed.\\
$^\S$ Checkpoint selected on the held-out evaluation set (optimistic); the final
checkpoint is reported in Table~\ref{tab:dx}.
}
\end{table}

\subsection{Passing the gates does not make outcome-only GRPO safe}
\label{subsec:plain-shortcuts}

Diagnostics provides the clearest test of whether RL can improve an adaptive
clinical procedure. The task passes both eligibility gates. Its best constant
gap is $+0.161$, which shows that gathering information can outperform the best
fixed action. The exact oracle requests $3.9$ findings per case in expectation over
the case distribution ($3.8$ on the held-out set); we use the expected value as the
reference oracle level in figures.
The base model also has nonzero support for oracle-grade trajectories under
stochastic decoding. The task therefore contains adaptive headroom, and the
initial policy can produce at least some behaviour that RL could reinforce.

Outcome-only GRPO does not reliably convert this opportunity into safer
information gathering. Across three training seeds, overall value increases from
$+0.251$ for the base model to $+0.285\pm0.027$. This aggregate improvement is
accompanied by deterioration on the safety-critical subgroup. High-risk value
falls from $+0.160$ to $+0.098\pm0.037$, while high-risk support falls from
$0.50$ to $0.33\pm0.29$.

The variance across seeds is informative, but it must be read against a very small
subgroup. The held-out diagnostics set contains only two high-risk cases, so
high-risk support of $0.50$ means that one of the two cases is reached by at least
one of twelve rollouts. The base model supports one abdominal case; one of the three
plain GRPO seeds loses it, falling to $0.0$. Mean questioning also falls from $2.5$
at initialization to $1.82\pm0.11$, well below the oracle level. The policy is
therefore improving its aggregate score partly by reducing the amount of evidence it
gathers. We treat the high-risk collapse as an illustrative failure rather than a
precise estimate; the moderate subgroup ($n{=}35$) and overall value carry the
statistically more informative evidence.

This result establishes the principal failure mode of outcome-only optimisation.
Passing the adaptive headroom and latent support gates identifies a setting in
which RL could help; it does not ensure that the reward will favour the desired
procedure. In diagnostics, outcome-only GRPO improves the mean while making rare
high-risk behaviour less reliable.

\subsection{Credit assignment improves diagnostics without seed collapse}
\label{subsec:credit-diagnostics}

Credit assignment changes the training signal but not the evaluation objective.
During optimisation, the model receives additional credit for questions that
reduce uncertainty about the hidden state. During held-out evaluation, it receives
only the original clinical utility minus question cost. An improvement therefore
requires the shaped policy to perform better under the same terminal verifier as
the control.

Across three seeds, credit assignment improves the principal diagnostics metrics.
Overall value increases from $+0.285\pm0.027$ with outcome-only GRPO to
$+0.313\pm0.024$. High-risk value increases from $+0.098\pm0.037$ to
$+0.157\pm0.057$. Moderate support rises from $0.38\pm0.18$ to
$0.67\pm0.10$. This moderate subgroup is particularly informative because
additional findings frequently change the preferred disposition; the improvement
is therefore consistent with more effective use of information rather than a
fixed action bias.

High-risk support is consistent with, but cannot establish, greater stability.
Credit-assignment GRPO maintains high-risk support of $0.50{\pm}0.00$ across all
three seeds, in each case the same single abdominal case that the base model
supports, whereas one plain GRPO seed loses that case. With two high-risk cases this
difference is suggestive only. The clearer gains are on the moderate subgroup
($n{=}35$), where support rises from $0.38{\pm}0.18$ to $0.67{\pm}0.10$, and on
overall value, where the improvement from $+0.285{\pm}0.027$ to $+0.313{\pm}0.024$
is modest and the seed ranges overlap.

Credit assignment does not, however, restore oracle-level gathering. Its policies ask
$2.08{\pm}0.30$ questions on average, compared with $1.82{\pm}0.11$ for plain GRPO,
$2.5$ for the base model, and an oracle level of approximately $3.9$; on the
high-risk cases they ask between $1.1$ and $1.5$. The improvement is therefore better
allocation of a still-shortened procedure rather than recovery of the full
gather-then-commit behaviour.

These results support a limited conclusion. Dense information credit improves
overall value and moderate support under the unshaped verifier and does not show the
high-risk seed failure observed under plain GRPO, but the high-risk evidence rests on
two cases and three seeds. A larger stratified evaluation is needed before treating
the high-risk stabilization as established.

\begin{table}[ht]
\centering

\caption{Diagnostics results in the held-out hidden condition using the pure
verifiable reward. Plain GRPO and credit-assignment GRPO are reported as
mean$\pm$sample standard deviation over three training seeds. Base and decoupled
GRPO are single-seed reference policies. Credit assignment improves overall
value, high-risk value, and moderate oracle-grade support relative to the outcome-only control. One plain GRPO seed loses the supported high-risk case, whereas
credit assignment retains it in each tested seed. ``supp'' denotes oracle-grade support@12 and should be
interpreted as sampled capability rather than single-trajectory reliability.
The held-out high-risk subgroup contains only two cases; every policy with high-risk
support of $0.50$ supports the same abdominal case, so high-risk values in this table
are illustrative rather than precise estimates. $^\S$The step-40 checkpoint was
selected on the held-out evaluation set and is therefore an optimistic estimate; the
final checkpoint is shown for comparison.}

\label{tab:dx}

\resizebox{\textwidth}{!}{%
\begin{tabular}{lccccc}
\toprule
Policy & overall value & high-value & high supp & moderate supp & mean asks \\
\midrule
base (untrained) & $+0.251$ & $+0.160$ & $0.50$ & $0.26$ & $2.5$ \\
plain GRPO (control) & $+0.285{\pm}0.027$ & $+0.098{\pm}0.037$ & $0.33{\pm}0.29$ & $0.38{\pm}0.18$ & $1.82{\pm}0.11$ \\
credit-assignment GRPO ($\lambda{=}0.2$) & $\mathbf{+0.313{\pm}0.024}$ & $+0.157{\pm}0.057$ & $\mathbf{0.50{\pm}0.00}$ & $\mathbf{0.67{\pm}0.10}$ & $2.08{\pm}0.30$ \\
decoupled GRPO (step 40$^\S$) & $+0.329$ & $\mathbf{+0.248}$ & $\mathbf{1.00}$ & $0.63$ & $4.4$ \\
decoupled GRPO (final, step 60) & $+0.304$ & $+0.158$ & $0.50$ & $0.74$ & $5.0$ \\
\bottomrule
\end{tabular}%
}
\end{table}

\begin{figure}[ht]
\centering
\includegraphics[width=\textwidth]{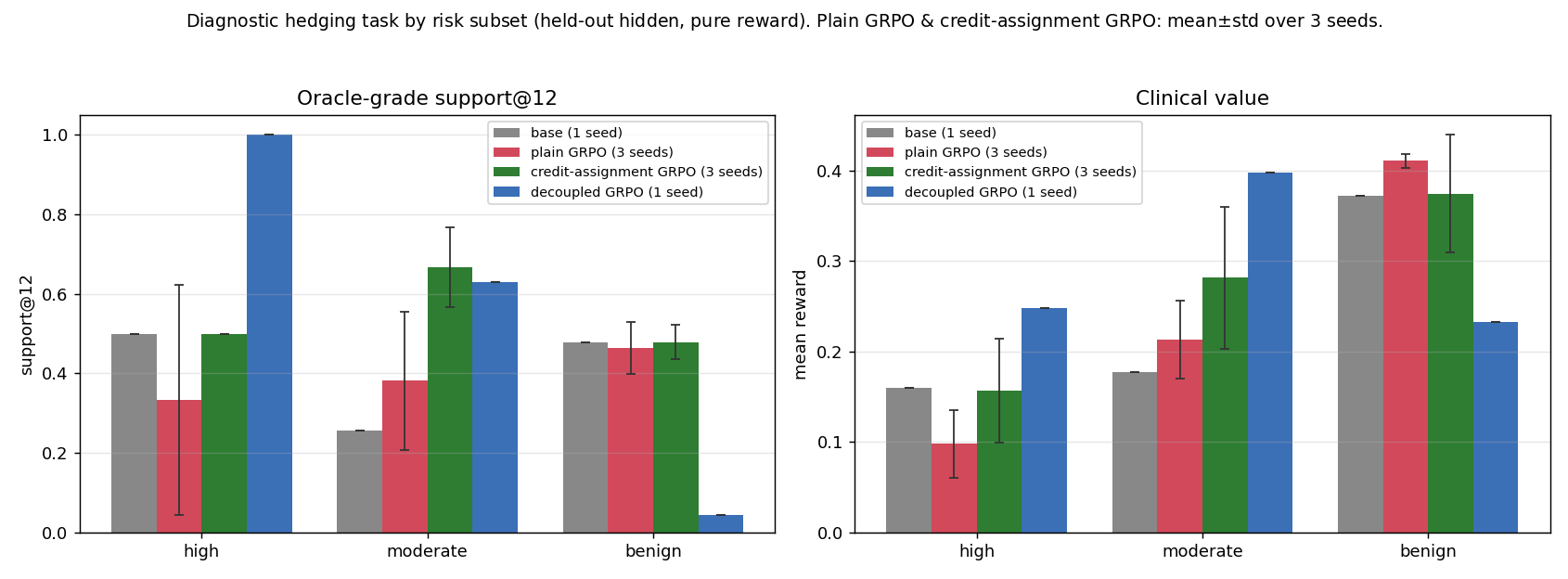}
\caption{Diagnostics performance by risk subgroup in the held-out hidden
condition. Every method is evaluated with the pure verifier reward. Plain GRPO
and credit-assignment GRPO show mean$\pm$standard deviation over three training
seeds; base and decoupled GRPO are single-seed references. Outcome-only GRPO
raises aggregate value but reduces and destabilizes high-risk performance.
Credit assignment improves the moderate subgroup and maintains high-risk support
of $0.50$ (one of two high-risk cases) across all three seeds. Decoupled GRPO produces the highest high-risk
value and support, but requests substantially more information.}
\label{fig:dx4arm}
\end{figure}

\FloatBarrier
\subsection{Decoupling increases high-risk support but also question burden}
\label{subsec:decoupled-results}

Credit assignment still requires one policy trajectory to balance two competing
objectives. Asking an additional question may improve information credit, while
committing earlier avoids the explicit question cost. We therefore test a
decoupled exploration and exploitation policy in which gathering and commitment
receive separate updates.

At its best checkpoint, decoupled GRPO produces the strongest high-risk result in
diagnostics. Relative to credit-assignment GRPO, high-risk value increases from
$+0.157$ to $+0.248$, and high-risk support increases from $0.50$ to $1.00$.
Overall value reaches $+0.329$.

This checkpoint was selected on the held-out set, so it is an optimistic estimate of
the method. By the final checkpoint, high-risk value returns to $+0.158$ and
high-risk support to $0.50$, matching credit-assignment GRPO, while mean questions
rise further to $5.0$. The high-risk advantage of decoupling is therefore transient
in this run, whereas its over-gathering persists. The full checkpoint trajectory is
shown in Figure~\ref{fig:dreamtraj} (Appendix~\ref{app:addendum}).

The gain is accompanied by greater interaction burden. Mean questions rise to
$4.4$, above the base model, the credit-assignment policy, and the oracle level
of approximately $3.9$. The increase is not distributed uniformly across
clinical severity. In the checkpoint analysis, benign support falls from $0.48$
to $0.04$, indicating that the explorer continues requesting information in
cases where earlier commitment would have been preferable.

Decoupled GRPO is therefore not a uniformly superior method. It moves the policy
toward a conservative operating point that prioritizes high-risk coverage at the
cost of additional questioning and weaker low-acuity efficiency. Such a tradeoff
could be reasonable when a missed emergency has substantially greater cost than
an unnecessary question, but the present experiment does not establish the
clinical utility of that operating point. The result is also based on one training
seed and should be interpreted as exploratory.

\FloatBarrier
\subsection{Information credit is a task-specific control parameter}
\label{subsec:titration-results}

The titration task tests whether the information credit coefficient transfers
across decision structures. The agent chooses among no therapy, reduced intensity,
and full intensity while balancing clot and bleed risk. The task passes both gates,
but its adaptive headroom is limited. The best constant gap is only $+0.075$.
The exact oracle uses all three terminal actions; gathering changes the preferred
action in $85\%$ of cases, and the optimal priority changes between clot-first and
bleed-first in $64\%$ of cases. Base support@12 is $0.52$ overall and $0.63$ on
the relevant subgroup.

The coefficient $\lambda$ controls how strongly an informative question is valued
relative to its cost. With no information credit, plain GRPO under-gathers. Mean
questions fall from $2.76$ for the base model to $1.40$, and the one-question
rate rises to $0.91$. Although high clot support reaches $1.00$, trajectory
inspection shows that this result is produced by an aggressive low-information
policy rather than adaptive risk assessment. The guarded report therefore labels
the behaviour \textsc{exploit}.

A weak coefficient, $\lambda{=}0.03$, does not correct the collapse. Mean
questions fall further to $1.24$, and the one-question rate reaches $0.93$. A
strong coefficient, $\lambda{=}0.2$, produces the opposite failure. Mean
questions increase to $4.10$, substantially above the oracle level ($2.77$
questions in expectation over the case distribution; $3.0$ on the held-out set), while overall value falls to $-0.091$.

The intermediate setting, $\lambda{=}0.1$, produces the most calibrated gathering
depth. This value follows the approximate relationship
$\lambda \cdot \Delta H_{\text{ask}} \gtrsim c$, where a useful question reduces
posterior entropy by approximately $0.33$ nats and the question cost is $0.03$.
At $\lambda{=}0.1$, mean questions are $2.67$, close to the oracle level. The one-question rate falls to $0.62$, and high clot support reaches $1.00$ without the
same degree of under-gathering.

Calibrated gathering does not produce the highest overall value. At
$\lambda{=}0.1$, overall value is $-0.036$, compared with $-0.022$ for the base
model and $-0.012$ for plain GRPO. This result is consistent with the small best
constant gap. The coefficient can improve the shape of the interaction, but a task
with little adaptive headroom offers limited opportunity to convert better
gathering into a higher terminal value.

\begin{table}[ht]
\centering
\caption{Titration results in the held-out hidden condition using the pure
verifier reward. The information credit coefficient controls the depth of
gathering. Plain GRPO and $\lambda{=}0.03$ produce high one-question rates and
under-gathering. At $\lambda{=}0.2$, the policy asks more than the oracle and
loses aggregate value. The analytically motivated setting
$\lambda{=}0.1$ places mean questioning close to the oracle level and preserves
high clot support. The absence of an overall value gain is consistent with the
small best constant gap of $+0.075$, which leaves limited adaptive headroom.}
\label{tab:ti}

\resizebox{\textwidth}{!}{%
\begin{tabular}{lccccc}
\toprule
Policy & overall value & overall asks & one-ask rate & high-clot value & high-clot supp \\
\midrule
base (untrained)                 & $-0.022$ & $2.76$ & $0.74$ & $+0.396$ & $0.67$ \\
plain GRPO (control)             & $-0.012$ & $1.40$ & $0.91$ & $+0.627$ & $1.00$ \\
\,+ credit-assignment ($\lambda{=}0.03$) & $-0.010$ & $1.24$ & $0.93$ & $+0.825$ & $1.00$ \\
\,+ credit-assignment ($\lambda{=}0.1$)  & $-0.036$ & $2.67$ & $0.62$ & $+0.556$ & $1.00$ \\
\,+ credit-assignment ($\lambda{=}0.2$)  & $-0.091$ & $4.10$ & $0.31$ & $+0.446$ & $0.67$ \\
\bottomrule
\end{tabular}%
}
\end{table}

\begin{figure}[ht]
\centering
\includegraphics[width=0.7\textwidth]{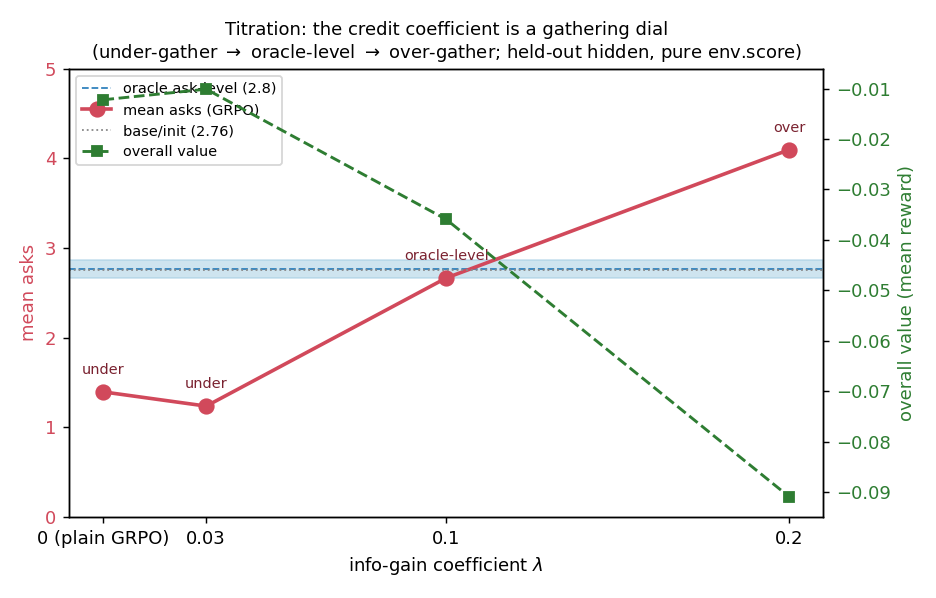}
\caption{Information credit coefficient sweep on the titration task. Mean
questions remain below the oracle level when $\lambda{\le}0.03$, approach the
oracle at $\lambda{=}0.1$, and exceed it at $\lambda{=}0.2$. The coefficient
therefore behaves as a task-specific control parameter for gathering depth,
rather than as a constant that can be transferred without calibration.}
\label{fig:tidial}
\end{figure}

\FloatBarrier
\subsection{A strong SFT warmup leaves little safe RL headroom}
\label{subsec:bridge-result}

Refill triage evaluates RL after a supervised protocol warmup. The question is no
longer whether reinforcement learning can discover the task from the base model.
It is whether RL improves a supervised policy that already demonstrates much of
the desired medication safety procedure.

The SFT warmup reaches high-risk value of $+0.195$ and high-risk support of
$1.00$, with mean questioning of $2.8$. Final plain GRPO and credit-assignment GRPO also show high-risk support of $1.00$. This apparent preservation is
misleading. The guarded report shows that support is generated by stochastic
sampling around policies that gather substantially less information. Mean
questions fall to $1.3$ under plain GRPO and $1.6$ under credit assignment.
High-risk value falls to $+0.032$ and $+0.043$, respectively.

The checkpoint trajectory reveals a short-lived difference between the two RL
methods. At the comparable early checkpoint, credit assignment maintains high-risk value close to the SFT warmup, at $+0.205$, whereas the plain GRPO control
has already fallen to $-0.007$. Continued optimisation removes this advantage,
and both final policies converge toward cheaper under-gathering procedures.

The refill experiment reproduces the diagnostics pattern from a supervised starting
point. The SFT warmup buys its high-risk performance with additional questioning:
overall value falls below the base model ($+0.318$ versus $+0.386$), largely because
benign cases are over-investigated. RL then recovers and exceeds the base model's
aggregate value ($+0.431$ for plain GRPO, $+0.415$ for credit assignment) by trimming
questions, while high-risk value falls below even the untrained base model ($+0.032$
and $+0.043$ versus $+0.151$). Reducing the number of questions was the easiest way to
raise the terminal reward, and the high-risk procedure the warmup had learned was the
cost.

An apparent gain from RL beyond SFT should therefore be audited by subgroup. In this
experiment, RL improved aggregate reward after SFT while eroding the high-risk
behaviour that SFT had established; the remaining reward pressure favoured efficiency
over the learned safety procedure.

\FloatBarrier
\subsection{Latent support does not guarantee repair of biased SFT}
\label{subsec:coverage}

The final experiment tests whether RL can repair a rare coverage gap inherited
from supervised training. We construct a diagnostics SFT dataset that under-represents high-risk emergency trajectories. The resulting policy performs well
on aggregate metrics while retaining a clinically important weakness.

The biased SFT reaches overall value of $+0.337$ and overall support of $0.93$,
but its high-risk value is $-0.285$. The discrepancy illustrates how aggregate
performance can conceal a rare failure. Under severe bias, in which all high-risk
examples are removed, high-risk support is $0.00$ and the model never generates
\texttt{emergent\_tx}. Gate 2 therefore fails. Under milder bias, in which
$30\%$ of high-risk examples remain, the model generates
\texttt{emergent\_tx} in $21\%$ of sampled trajectories. Gate 2 passes, which
shows that the missing action remains present in the policy distribution.

The mildly biased SFT policy, which serves as the initialization for repair, has a
reliability gap rather than an availability gap. Its high-risk support is $1.00$: for
every high-risk case, at least one of twelve rollouts is oracle-grade. Its typical
high-risk rollout is nevertheless poor, with high-risk value of $-0.285$.

Both RL methods change this profile without repairing it. Credit-assignment GRPO
raises high-risk value to $+0.092$ but reduces high-risk support from $1.00$ to
$0.14$; mean questions fall from $3.0$ to $2.2$ overall and to $1.4$ on high-risk
cases, and overall support falls from $0.93$ to $0.45$. The policy avoids the worst
outcomes more often but rarely reaches the oracle-grade procedure that the
initialization could still sample.

Decoupled GRPO moves in the opposite direction on gathering. Mean questions rise to
$4.3$, high-risk value reaches $+0.040$, and high-risk support falls to $0.36$. Benign
support falls from $0.92$ to $0.16$, and overall support falls to $0.41$. Neither
method converts the sampled oracle-grade behaviour into reliable behaviour
(Figure~\ref{fig:reptraj}, Appendix~\ref{app:addendum}).

The experiment clarifies the role of the latent support gate. If the initial
policy never samples the desired action, RL has no successful trajectory to
amplify. Nonzero sampling is therefore a necessary condition for repair. It is
not sufficient. When the desired behaviour is rare within an otherwise strong
policy, GRPO may reinforce a more accessible neighbouring strategy, such as
asking less or asking broadly, rather than selectively restoring the missing
trajectory.

\begin{table}[ht]
\centering
\caption{Attempted repair of a mildly biased diagnostics SFT policy ($30\%$ of
high-risk SFT examples retained), in the held-out hidden condition with $n{=}200$
cases ($14$ high-risk) and the pure verifier reward. The SFT initialization samples
an oracle-grade trajectory for every high-risk case (support $1.00$) but performs
poorly on a typical high-risk rollout (value $-0.285$). Both RL methods raise
high-risk value but sharply reduce high-risk support, and both reduce overall
support. Credit-assignment GRPO under-gathers; decoupled GRPO over-gathers and
damages benign support. Neither converts the sampled behaviour into a reliable
procedure. ``supp'' denotes oracle-grade support@12.}
\label{tab:repair}

\resizebox{\textwidth}{!}{%
\begin{tabular}{lcccccc}
\toprule
Policy & overall val & overall supp & high val & high supp & benign supp & mean asks \\
\midrule
biased SFT (init)     & $+0.337$ & $\mathbf{0.93}$ & $-0.285$ & $\mathbf{1.00}$ & $\mathbf{0.92}$ & $3.0$ \\
\,+ credit-assignment GRPO    & $+0.236$ & $0.45$ & $+0.092$ & $0.14$ & $0.45$ & $2.2$ \\
\,+ decoupled-GRPO             & $+0.292$ & $0.41$ & $+0.040$ & $0.36$ & $0.16$ & $4.3$ \\
\bottomrule
\end{tabular}%
}

\end{table}

\begin{figure}[ht]
\centering
\includegraphics[width=0.92\textwidth]{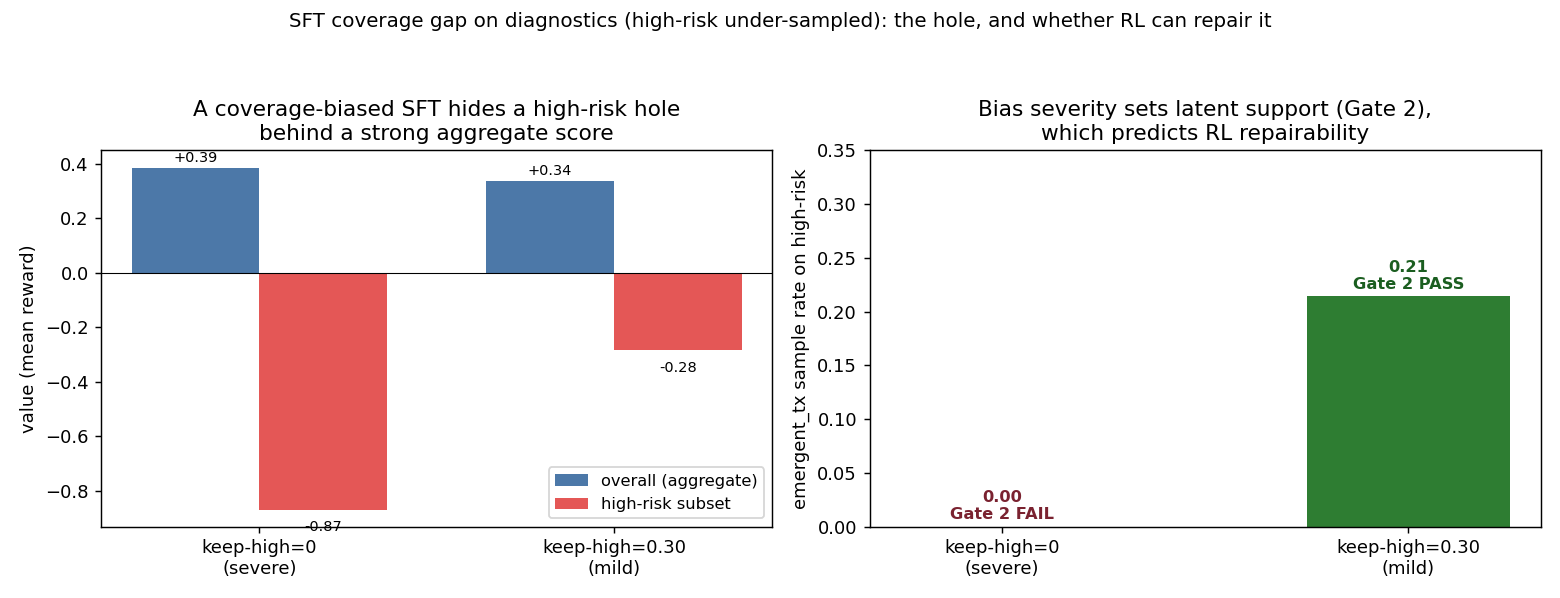}
\caption{Effect of supervised coverage bias on aggregate performance and latent
support. Left, the biased SFT retains high overall support despite poor high-risk
value. Right, severe bias removes the high-risk action
\texttt{emergent\_tx} from sampled trajectories and fails Gate 2. Mild bias
retains a $21\%$ sampling rate and passes Gate 2. The subsequent repair
experiment shows that nonzero latent support is necessary but not sufficient for
RL to restore the missing behaviour.}
\label{fig:hole}
\end{figure}

\FloatBarrier
\subsection{Summary of the empirical findings}

Across the experiments, no single RL method dominates across clinical severity,
interaction burden, and task structure. Outcome-only GRPO frequently improves
aggregate value by shortening the dialogue, but this can reduce high-risk
reliability. Credit assignment produces the strongest replicated improvement in
diagnostics; it raises held-out value and removes the observed high-risk seed
collapse. Decoupled exploration further improves high-risk support, but imposes
additional burden on lower-acuity cases. In titration, the benefit of information
credit depends on its scale and on the limited adaptive headroom of the task.

The SFT results place a further boundary on the value of RL. When supervised
training already captures the desired procedure, RL may have little safe
improvement left to offer and can instead erode the initialization. When SFT
contains a rare coverage gap, the ability to sample the missing action does not
ensure that RL will repair it.

The main empirical conclusion is therefore conditional. Reinforcement learning
can improve clinical information gathering when the task contains adaptive
headroom, the initial policy exposes useful trajectories, and the reward assigns
credit to the evidence gathering process. When these conditions are absent or
poorly calibrated, RL can replace one failure mode with another while preserving
or even improving the aggregate score.

\FloatBarrier
\section{Discussion}
\label{sec:discussion}

The central conclusion is not that reinforcement learning consistently improves
clinical multi-turn agents. RL can improve, preserve, or degrade the decision
procedure depending on the task, the initial policy, and the structure of the
reward.

These findings suggest that the important question is not simply whether RL
raises average reward. The more useful question is whether RL teaches a reliable
procedure for gathering evidence and committing to an action. In the present
environments, those two outcomes are not equivalent. A policy can improve its
mean score by asking fewer questions, choosing a broadly aggressive action, or
using an intermediate decision that performs adequately on common cases. Such a
policy may optimise the verifier without learning the clinical process that the
verifier was intended to encourage.

\paragraph{A practical decision rule for RL post-training}
The experiments support a staged approach to reinforcement learning for clinical
agents. First, the task should be tested for adaptive headroom. When the best
fixed action performs nearly as well as the exact adaptive oracle, there is little
decision value for RL to recover through information gathering. Second, the
initial policy should be tested for latent behavioural support. If it never
samples the desired gather-then-commit trajectory, policy optimisation has no
successful behaviour to amplify. Third, when both gates pass, the reward should
assign credit to clinically useful intermediate decisions rather than relying
only on the final action.

These gates are negative screens, not certificates of safety. Passing them shows
that RL has a plausible opportunity to help; it does not show that outcome-only
GRPO will use that opportunity appropriately. Diagnostics passes both gates, yet
plain GRPO improves aggregate value while reducing high-risk performance and
question depth. The reward structure remains the controllable factor.

Supervised initialization adds a fourth consideration. If SFT has already learned
most of the behaviour measured by the verifier, RL may have little safe
headroom. If the SFT policy contains a rare coverage gap, the presence of the
missing action in occasional stochastic rollouts does not ensure that standard RL
will repair the associated procedure. A practical pipeline should therefore
measure both SFT performance and residual error structure before deciding whether
additional RL is justified.

\paragraph{The safety question concerns procedure, not only outcome}
Clinical decisions are usually justified by evidence. A medication refill may be
appropriate only after checking renal function, contraindications, or adverse
effects. Emergency treatment may be appropriate only after obtaining findings
that support a high-acuity state. Two agents can select the same final action while
differing substantially in reliability; one may have gathered the necessary
evidence, while the other may have guessed correctly or followed a fixed action
bias.

This distinction explains why average reward and final action accuracy are
insufficient. In diagnostics, plain GRPO improves overall value relative to the
base model, but high-risk value and support decline. In refill triage, support at
twelve sampled rollouts remains high after RL, yet the typical policy asks far
fewer questions than the supervised initialization. The guarded analysis therefore
classifies the apparent support as an artefact of stochastic sampling around an
under-gathering policy, rather than evidence of a reliable clinical procedure.

Clinical evaluation should consequently include both outcome and process measures.
Relevant audits include whether the model obtained the findings needed to justify
its decision, whether it committed prematurely, whether it repeatedly used a
convenient intermediate action, whether it escalated without support, and whether
it gathered excessive information on benign cases. Rare high-risk cases should be
reported separately because aggregate metrics can remain stable while performance
on the most consequential subgroup deteriorates.

\paragraph{Implications for clinical decision support}
The present environments are not deployment studies, but the failure modes have
direct relevance to future clinical decision support systems. A model used for
refill review could reduce interaction burden by asking fewer questions while
silently omitting a renal or contraindication check. An acute assessment agent
could appear sensitive to emergencies because it escalates broadly, even when its
decisions are unsupported by the available evidence. A system designed to avoid
those misses could fail in the opposite direction by requesting every available
finding and creating unnecessary delay or workload.

These behaviours matter because clinical safety is not defined by sensitivity
alone. A useful system must balance missed danger, unnecessary escalation,
information burden, and the strength of evidence supporting its recommendation.
The results show that reward design can move the policy among these operating
points without producing the intended balance. A higher verifier score should
therefore not be interpreted as a clinically meaningful improvement unless the
trajectory analysis shows that the model gathered and used the appropriate
evidence.

The findings also have implications for supervision. A clinician reviewing only
the final recommendation may not detect that the policy reached it through an
unstable procedure. Supervision tools should expose which findings were requested,
which evidence changed the action, what remained unknown at commitment, and
whether the decision would have changed under plausible alternate findings. Such
process visibility may be as important as the final confidence score.

\paragraph{Why clinical information gathering differs from math and code RLVR}
RLVR has produced strong results in domains where correctness can be sharply
verified, including mathematics and programming
\citep{shao2024deepseekmath}. In those settings, an incorrect answer or failing
program usually receives an immediate terminal penalty. Clinical information
gathering contains an additional layer. The desired output is not only an action;
it is a sequence of decisions under uncertainty about which evidence to obtain,
how much evidence is sufficient, and when the cost of further investigation
exceeds its expected value.

This structure creates more opportunities for shortcut optimisation. A policy may
avoid the difficult high-risk decision by selecting an intermediate action. It may
ask one highly predictive question and ignore the remaining evidence needed for a
robust decision. It may also stop before obtaining information that would expose
an error in its preferred action. These procedures can receive acceptable average
reward even though they are not reliable clinical strategies.

The implication is not that RLVR is unsuitable for clinical agents. Rather, the
verifiable objective must represent more of the procedure. Credit assignment is
one approach because it rewards questions that reduce uncertainty about the hidden
state. Decoupling exploration from exploitation is another because it gives
information gathering its own optimisation signal. The tradeoffs observed here
show that neither mechanism is automatically safe; each changes the balance
between gathering, commitment, and burden.

\paragraph{SFT changes the interpretation of RL gains}
Most practical clinical models will undergo supervised adaptation before
reinforcement learning. The incremental question is therefore whether RL improves
what SFT leaves unresolved, rather than whether it can teach the task from an
unadapted base model.

The refill experiment illustrates the limited headroom case. The SFT warmup
already learns much of the medication safety procedure and reaches strong high-risk support. Continued RL raises aggregate reward, but it does so by moving toward
fewer questions and lower high-risk value. This does not establish that RL is
generally ineffective; it indicates that the supervised model had already captured
much of the safety-relevant benefit and that the remaining reward pressure favoured a
cheaper procedure.

The biased diagnostics experiment illustrates the opposite case. A supervised
policy can perform strongly on average while retaining a rare high-risk weakness.
RL is an appealing repair mechanism, but repair is not automatic. When the high-risk action is absent from the policy distribution, there is no useful trajectory
to reinforce. When the action is present but rare, both tested RL methods move
toward easier neighbouring strategies rather than selectively restoring the
missing behaviour.

These results suggest that supervised coverage remains a primary safety concern.
RL can amplify behaviour that is already represented and rewarded frequently
enough, but it should not be assumed to discover every omitted clinical pattern.
Targeted supervised examples, risk-balanced sampling, explicit subgroup rewards,
or curriculum design may be required before RL can provide a reliable incremental
gain.

\paragraph{Relation to clinical RLHF and preference optimisation}
The experiments use analytic verifiers because they make the state, utility, and
optimal adaptive policy measurable. This does not imply that every clinically
important objective can be reduced to a deterministic reward. Communication
quality, calibrated uncertainty, empathy, shared decision making, and clinician or
patient preferences often require human judgement. RLHF, direct preference
optimisation, and related methods are more natural for these dimensions.

The cautionary result nevertheless extends beyond analytic RLVR. A policy that can
exploit a deterministic verifier may also exploit a learned reward model. Learned
feedback can introduce additional vulnerabilities, including inconsistency,
supervisor bias, excessive caution, or preference for stylistically reassuring
responses over well-supported decisions. Process-level and subgroup evaluation
remain necessary regardless of whether the reward is analytic or learned.

A future clinical alignment pipeline will probably be hybrid. Deterministic
verification may be appropriate for components such as dose calculations,
contraindication checks, interaction checks, monitoring requirements, schema
validity, and minimum information sufficiency. Preference-based methods may be
better suited to communication, uncertainty expression, and preference sensitive
tradeoffs. Both components should be evaluated using the same guarded principle:
the final response is not sufficient evidence that the underlying procedure is
reliable.

\paragraph{Limitations}
The experiments use synthetic, stylized environments with deterministic patient
responses and hand-specified utility matrices. They are designed to isolate
learning dynamics, not to estimate real clinical benefit or harm. The cases do
not include noisy histories, ambiguous language, documentation errors, uncertain
allergy records, conflicting laboratory results, changing patient goals, or
clinician disagreement. The utility matrices encode experimental assumptions
about relative consequences and have not been validated by clinical experts.
The trained policies should therefore not be interpreted as suitable for patient
care.

The evidence is also limited to one model family and scale,
\texttt{medgemma-4b-it}, with bounded LoRA-based training runs. The main
diagnostics comparison between outcome-only GRPO and credit assignment is
replicated across three seeds. The decoupled method, titration coefficient sweep,
refill experiments, and SFT repair experiments remain single-seed. Conclusions
from those cells are therefore exploratory, particularly where checkpoint
selection or stochastic support metrics are involved.

Support@$12$ is useful for measuring whether a behaviour exists within the policy
distribution, but it is not a direct estimate of single-trajectory deployment
safety. A policy can obtain high support because one of twelve samples succeeds
while the majority follow an unsafe procedure. Future studies should place greater
emphasis on deterministic or greedy evaluation, pass at one performance,
premature commitment rates, supported versus unsupported escalation, expected
regret, and subgroup-specific uncertainty intervals.

The SFT analysis is also incomplete. We evaluate one light protocol warmup and one
family of biased diagnostics initializations, but we do not measure a complete dose
curve across SFT strength and coverage. We therefore show that strong SFT can leave
little RL headroom and that RL can fail to repair a biased policy; we do not
characterize every supervised regime or every possible repair method.

Finally, the number of high-risk cases is small in the principal held-out sets (two
of 60 diagnostics cases), and the decoupled result relies on a checkpoint selected on
held-out data.
Reporting the subgroup separately prevents it from being hidden by the aggregate,
but larger stratified evaluations would produce more precise estimates.

An ongoing extension of this study addresses several of these limitations; we
outline it under Future work. The conclusions reported here rest only on the
experiments presented in this paper.

\paragraph{Future work}
An ongoing extension of this study investigates whether these findings generalize
to a larger model, an external interactive clinical benchmark, and
clinician-evaluated trajectories. It will repeat the principal SFT, RL, and
reward-design comparisons with MedGemma-27B, and evaluate trained policies on an
independent multi-turn clinical benchmark, such as AgentClinic
\citep{schmidgall2026agentclinic}, to test generalization beyond our
deterministic synthetic environments. A qualified clinician will review sampled
trajectories blind to training condition, assessing information gathering,
evidence sufficiency, premature commitment, unnecessary questioning, and
unsupported or unsafe escalation, as well as the clinical plausibility of the
hand-specified utilities. The extension will also enlarge the risk-stratified
evaluation sets, particularly the diagnostics high-risk subgroup, which here
contains only two cases, and report subgroup-specific results with uncertainty
estimates. Finally, it will select RL checkpoints on a separate validation split
and reserve an untouched test set for final evaluation, avoiding the optimistic
selection of the decoupled step-40 checkpoint. These experiments test the
generality of the present results; neither external benchmarking nor clinician
review would by itself establish clinical safety.

Beyond this extension, several broader questions remain open for us and the wider
research community. First, RL headroom
and repairability have not been characterized as functions of SFT strength and
coverage. A controlled study could train models on increasing amounts of
well-covered supervised data and evaluate them before and after matched RL, while
separately varying high-risk coverage with total SFT quality held approximately
constant. This would identify when RL has genuine residual headroom, when it can
repair a rare weakness, and when it merely reshapes an already strong policy.

Second, entropy reduction is useful because it is analytic, but it rewards
uncertainty reduction rather than the decision value of a question. Alternative
rewards could estimate expected regret reduction, the probability of changing the
optimal action, or satisfaction of explicit evidence-sufficiency requirements.
Such rewards may better distinguish clinically decisive questions from findings
that reduce uncertainty without affecting management.

Third, external benchmarks test transfer, but controlled environments remain
necessary when exact oracles and verifiers are required. A complementary direction
is to make such environments more realistic while retaining these properties, for
example through stochastic patient responses, ambiguous findings, conflicting
chart information, missing laboratory values, uncertain medication histories, and
multiple clinically acceptable stopping points. Expert-reviewed utilities and
evidence requirements would be needed before such environments could support
stronger clinical claims.

Fourth, the same patient state may call for a different operating point depending
on whether the objective is to minimise missed emergencies, reduce unnecessary
escalation, limit patient burden, or prepare a clinician handoff. Goal-conditioned
policies that represent these objectives explicitly could learn calibrated
tradeoffs rather than relying on one fixed utility matrix.

The long-term goal is a reusable benchmark for clinical multi-turn agents in which
RL is evaluated as procedure learning. Such a benchmark should measure whether the
agent requests the findings that matter, stops when enough is known, commits to an
action supported by the available evidence, and preserves safety on rare
high-consequence cases. Aggregate reward should remain part of the evaluation, but
it should not be the sole criterion for selecting a clinical policy.

\paragraph{Conclusion}
The experiments show that verifiable outcomes alone do not ensure reliable
clinical information gathering. Outcome-only GRPO can improve average reward while
reducing evidence gathering and weakening high-risk behaviour. Dense credit
assignment provides a meaningful improvement in the replicated diagnostics
comparison, but its benefit depends on task headroom and reward calibration.
Stronger exploration can improve high-risk coverage while increasing burden, and
RL does not reliably add safe value beyond a strong SFT policy or repair rare omissions
in biased supervised data.

The broader implication is that reinforcement learning for clinical agents should
be treated as optimisation of a sequential decision procedure, not merely a final
answer. Safe post-training requires adaptive headroom, behavioural support in the
initial policy, appropriate intermediate credit, and evaluation that exposes rare
procedural failures rather than averaging them away.

\section*{AI-Assisted Research and Manuscript Preparation}

The authors used Anthropic Claude, including Claude Code,
and OpenAI ChatGPT to assist with software development,
debugging, experimental infrastructure, running experiments,
data analysis and figure generation, literature research,
manuscript drafting, and editorial revision.
AI-assisted outputs were reviewed and validated by the
human authors. The authors retain full responsibility for
the research design, experimental results, scientific
interpretation, and manuscript content.

\FloatBarrier
\appendix

\section{Implementation details}
\label{app:impl}

All experiments adapt the text-only \texttt{medgemma-4b-it} base model using
LoRA with rank $32$ and $\alpha{=}64$. Tables~\ref{tab:sfthp} and
\ref{tab:rlhp} report the supervised and reinforcement learning hyperparameters.

Training uses a disaggregated two-node system. One NVIDIA DGX Spark node performs
model optimisation, while a second node runs a vLLM server that generates
on policy trajectories. Both systems use Grace Blackwell hardware with unified
memory. A supervised warmup requires approximately $0.5$ hours, while a
$60$-step GRPO run generally requires $6$ to $8$ hours.

\begin{figure}[ht]
\centering
\includegraphics[width=\textwidth]{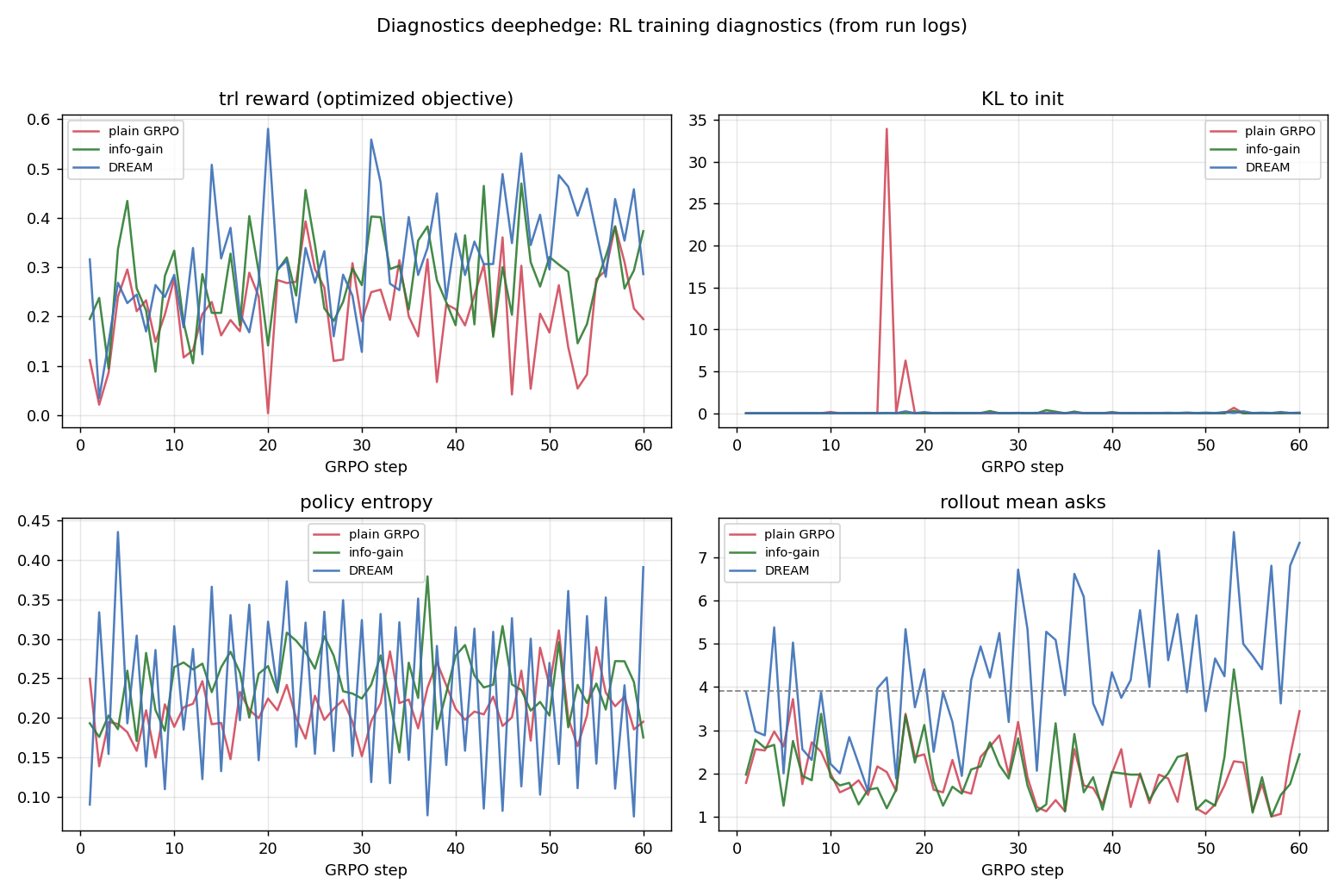}\\[4pt]
\includegraphics[width=0.5\textwidth]{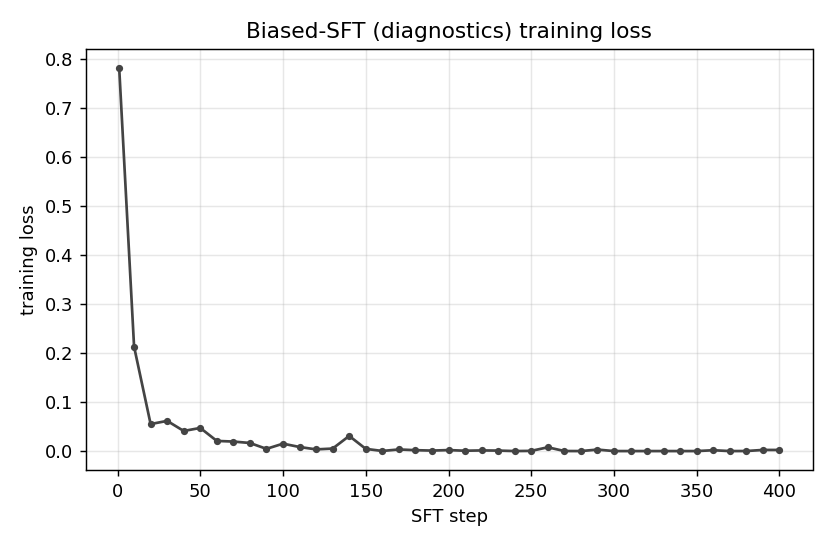}
\caption{Training diagnostics derived from the run logs. Top, diagnostics GRPO
over $60$ optimisation steps. The panels show the optimised reward, KL divergence
from the initialization, policy entropy, and mean questions per rollout for
outcome-only GRPO, credit-assignment GRPO, and decoupled GRPO. The question
trajectory reveals the characteristic procedural differences among the methods;
outcome-only GRPO and credit-assignment GRPO both settle well below the oracle
gathering level of approximately $3.9$, with credit assignment slightly higher,
while decoupled GRPO rises above it. Bottom, training loss for the biased SFT warmup. Loss converges to
approximately $0.002$, showing that the supervised trajectories are closely fit
before RL begins.}
\label{fig:traincurves}
\end{figure}

Figure~\ref{fig:traincurves} summarizes the principal optimisation diagnostics.
The upper panels show reward, divergence from the initialization, policy entropy,
and mean questions per rollout for the diagnostics experiments. The lower panel
shows the loss curve for the biased SFT warmup. All values are read directly from
the corresponding training logs.

\begin{table}[ht]
\centering
\caption{Hyperparameters for the supervised protocol warmup. Loss is computed
on every assistant turn; system messages, user messages, and simulator responses
are excluded from the supervised objective.}
\label{tab:sfthp}
\begin{tabular}{ll}
\toprule
Base model & \texttt{medgemma-4b-it} (text-only causal LM) \\
Adapter & LoRA $r{=}32$, $\alpha{=}64$, dropout $0.05$, all-linear targets \\
Objective & causal-LM cross-entropy on \emph{every} assistant turn (system/user masked) \\
Optimizer & AdamW, lr $1{\times}10^{-4}$, linear warmup $5\%$ then constant \\
Epochs / batch / max seq len & $2$ / $4$ / $2048$ \\
Training trajectories & $800$ (oracle-adaptive; generation seed $7$) \\
Seed / checkpoint & $1$ / \texttt{final} \\
\bottomrule
\end{tabular}
\end{table}

\begin{table}[ht]
\centering
\caption{Hyperparameters for GRPO training. Outcome-only GRPO uses the terminal
utility minus the accumulated question cost. Credit assignment adds
$\lambda\Phi$ to the rollout reward during training only. Decoupled GRPO instead
alternates exploration updates based on state recoverability with exploitation
updates based on \texttt{env.score}. Every policy is evaluated using the same
unshaped terminal verifier.}
\label{tab:rlhp}
\begin{tabular}{ll}
\toprule
Algorithm & GRPO (TRL), custom on-policy \texttt{rollout\_func} \\
Initialization & base model, or merged SFT adapter \\
Adapter & LoRA $r{=}32$, $\alpha{=}64$ (q,k,v,o,gate,up,down) \\
Group size (rollouts/prompt) & $8$ \\
Per-device batch / grad-accum & $8$ / $4$ \\
Training cases / steps / save & $300$ / $60$ / every $20$ \\
Learning rate / KL $\beta$ & $1{\times}10^{-5}$ / $0.02$ \\
Rollout temperature / max tokens-per-turn & $1.0$ / $128$ \\
Max completion length / turn cap & $512$ / $10$--$12$ \\
Importance-sampling correction & off (fully on-policy, ratio $\approx 1$) \\
EOS token ids / token masking & $\{1,106\}$ / \texttt{env\_mask} (model tokens only) \\
Reward & $U[\mu][a]-0.03\,n_{\text{asks}}$ ($+\lambda\Phi$ for credit assignment) \\
Credit assignment $\lambda$ & $0.2$ (diagnostics/refill); swept $\{0,0.03,0.1,0.2\}$ (titration) \\
Seed & $1$ (diagnostics plain/credit assignment: $1$--$3$) \\
\bottomrule
\end{tabular}
\end{table}

\FloatBarrier
\section{Representative oracle trajectories}
\label{app:cases}

The following examples illustrate the common ask-to-commit structure used across
the three environments. Each trace shows one held-out case and the corresponding
exact adaptive oracle trajectory. The simulator reveals only findings that the
agent explicitly requests; all other findings remain hidden.

These examples are intended to clarify the experimental interface and scoring
procedure. They are simplified simulator traces rather than realistic clinical
conversations.

\begin{verbatim}
DIAGNOSTICS (deephedge) -- true latent: high (emergency); oracle: emergent_tx
  user:      Presentation: adult with acute chest pain. Chart: no results yet.
  assistant: {"action_type":"ask","intent":"troponin_high"}   user: troponin elevated
  assistant: {"action_type":"ask","intent":"ecg_ischemia"}    user: ECG shows ischemia
  assistant: {"action_type":"ask","intent":"dyspnea_severe"}  user: severe dyspnea
  assistant: {"action_type":"ask","intent":"hypotension"}     user: hypotensive
  assistant: {"action_type":"commit","decision":"emergent_tx",...}   reward = 0.83

REFILL TRIAGE -- true latent: moderate; oracle: dose_adjust
  user:      Refill request: ACE inhibitor / ARB (lisinopril). Chart: no recent labs.
  assistant: {"action_type":"ask","intent":"renal_decline"}     user: eGFR declining
  assistant: {"action_type":"ask","intent":"egfr_severe"}       user: decline mild
  assistant: {"action_type":"ask","intent":"volume_dehydrated"} user: dehydrated
  assistant: {"action_type":"commit","decision":"dose_adjust",...}   reward = 0.81

TITRATION -- true latent: high clot / low bleed; oracle: high (full intensity)
  user:      Patient: older adult with atrial fibrillation. Chart: none documented.
  assistant: {"action_type":"ask","intent":"bleed_history"}  user: no prior bleeding
  assistant: {"action_type":"ask","intent":"fall_risk"}      user: low fall risk
  assistant: {"action_type":"commit","decision":"high",...}   reward = 1.34
\end{verbatim}

The diagnostics trajectory obtains four findings before selecting emergency
treatment. Its terminal utility is reduced by four question costs, producing the
reported reward of $0.83$. The refill trajectory obtains three findings before
modifying the medication dose, resulting in reward $0.81$. The titration
trajectory asks about two bleeding risk factors before selecting full treatment
intensity.

\FloatBarrier
\section{Additional decoupled, refill triage, and coverage results}
\label{app:addendum}

The following figures provide the decoupled GRPO checkpoint trajectory and
additional results for the refill triage and biased SFT experiments. They complement the main results by showing how
aggregate support can conceal deterioration in high-risk behaviour or excessive
information gathering.

\begin{figure}[ht]
\centering
\includegraphics[width=0.92\textwidth]{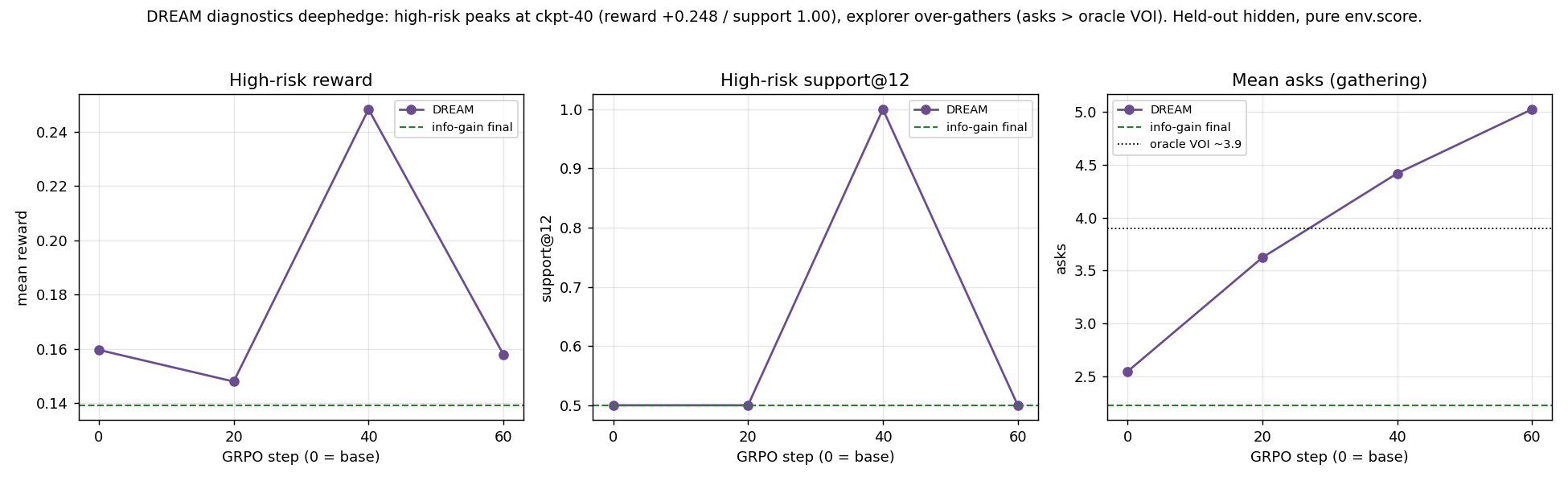}
\caption{Checkpoint trajectory for decoupled GRPO on diagnostics. Evaluation
uses the held-out hidden condition and the pure verifier reward. High-risk
oracle-grade support and high-risk value peak at checkpoint 40, reaching $1.00$
and $+0.248$, respectively. The same checkpoint also shows increased gathering;
mean questions approach and then exceed the oracle level of approximately $3.9$.
The trajectory illustrates the method's central tradeoff between high-risk
coverage and information burden.}
\label{fig:dreamtraj}
\end{figure}

\begin{figure}[ht]
\centering
\includegraphics[width=\textwidth]{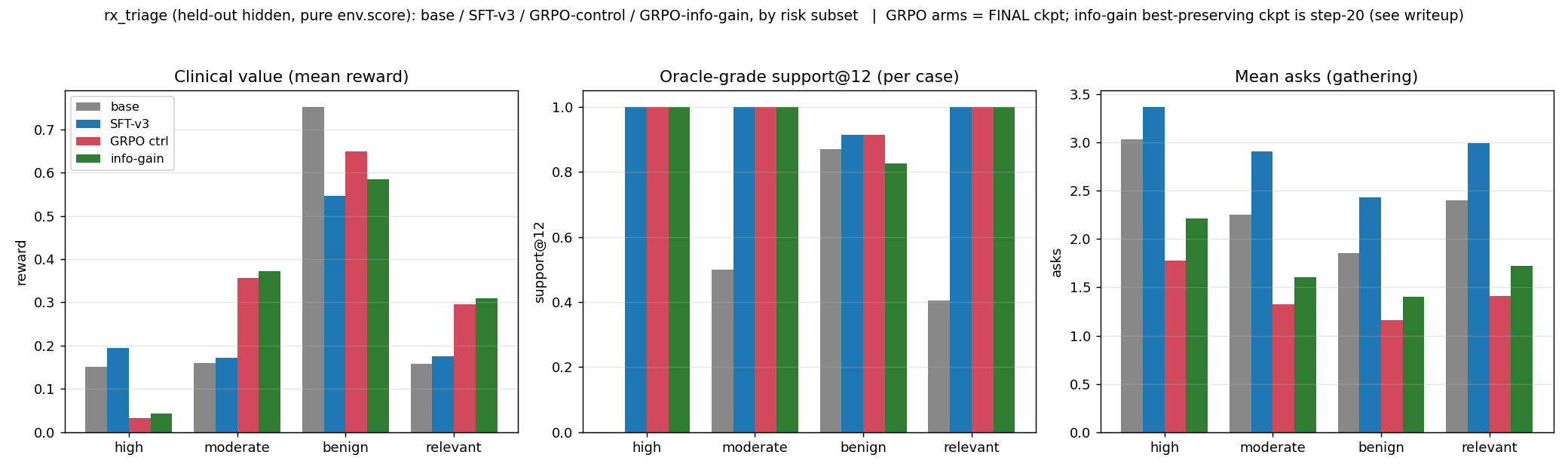}
\caption{Refill triage performance by risk subgroup in the held-out hidden
condition, evaluated with the pure verifier reward. The figure compares the base
model, the SFT v3 warmup, outcome-only GRPO, and credit-assignment GRPO at their
final checkpoints. The SFT warmup already achieves the strongest high-risk
result. Both RL policies subsequently reduce high-risk value and move toward
shorter, under-gathered trajectories, despite preserving high support under
stochastic sampling.}
\label{fig:rx4arm}
\end{figure}

\begin{figure}[ht]
\centering
\includegraphics[width=\textwidth]{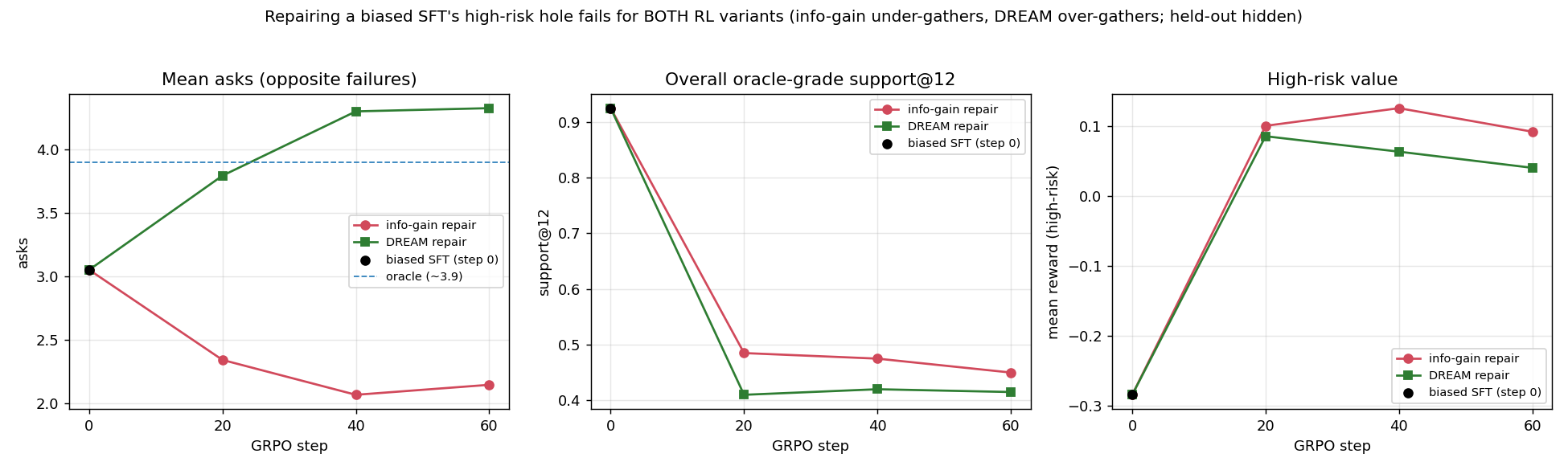}
\caption{Attempted reinforcement learning repair of the biased SFT policy.
Credit assignment and decoupled GRPO move away from the initialization in
opposite directions. Credit assignment reduces gathering and loses high-risk
support; decoupled GRPO increases gathering and damages benign support. Both
methods reduce the SFT policy's strong aggregate support, and neither restores
the missing high-risk procedure.}
\label{fig:reptraj}
\end{figure}

\begin{figure}[ht]
\centering
\includegraphics[width=\textwidth]{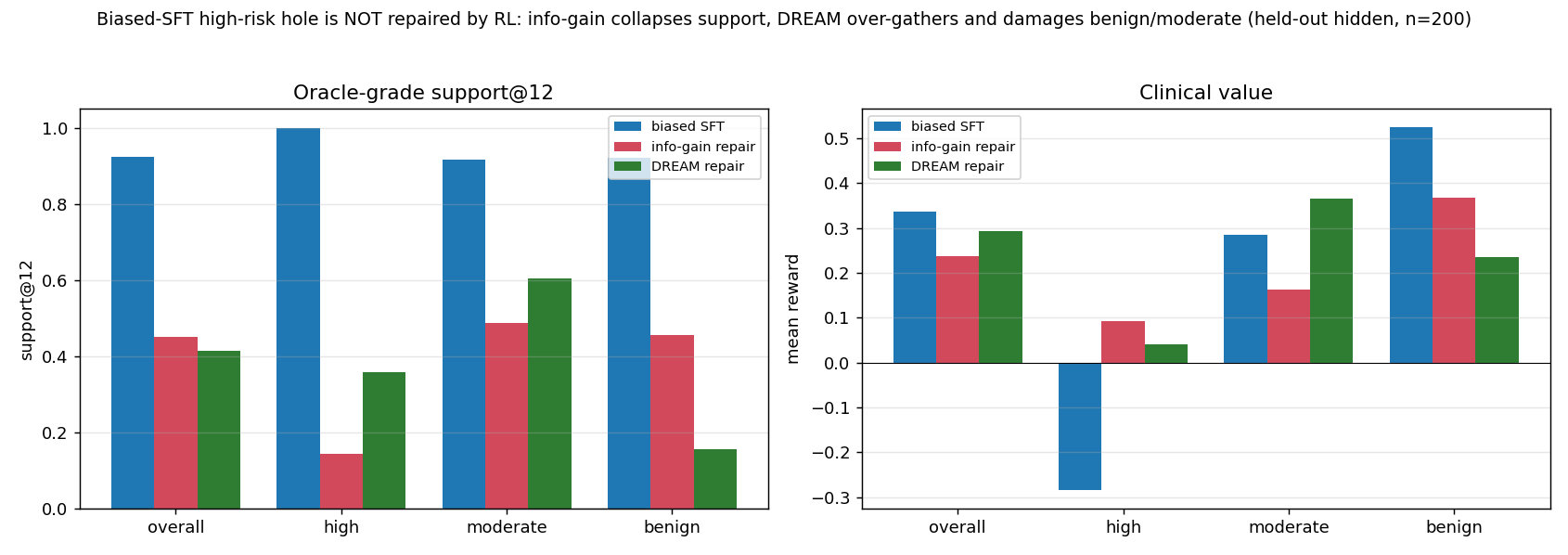}
\caption{Attempted repair of a mildly biased diagnostics SFT policy by risk
subgroup, using the held-out hidden condition and $n{=}200$ cases. Both RL methods
lose most of the initialization's high-risk oracle-grade support; support falls
from $1.00$ for the SFT initialization to $0.14$ with credit-assignment GRPO and
$0.36$ with decoupled GRPO, even as high-risk value rises. The methods fail in
different directions. Credit assignment reduces gathering, while decoupled GRPO
increases question burden and lowers benign support from $0.92$ to $0.16$. This
figure complements Table~\ref{tab:repair} and Figure~\ref{fig:reptraj}.}
\label{fig:rep4arm}
\end{figure}

\FloatBarrier
\section*{Code}
\url{https://github.com/csbrendan/Safe-Reinforcement-Learning-for-Multi-Turn-Clinical-Agents}

\label{sec:cite}
\bibliography{sample.bib}
\bibliographystyle{plainnat}

\clearpage
\end{document}